\pdfoutput=1
\documentclass[11pt]{article}
\usepackage[final]{acl}
\usepackage{times}
\usepackage{latexsym}
\usepackage[T1]{fontenc}
\usepackage[utf8]{inputenc}
\usepackage{fontawesome5}
\usepackage{microtype}
\usepackage{amsmath}
\usepackage{amssymb}
\usepackage{stfloats}
\usepackage{inconsolata}
\usepackage{graphicx}
\usepackage{amsmath} 
\usepackage{caption}
\usepackage{subcaption}
\usepackage{array}
\usepackage{booktabs}
\usepackage{multirow}
\title{Aligning Implied Statements for Implicit Hate Speech Generalizability with Context-Bounded Semi-hard Negative Mining}

\author{ Wicaksono Leksono Muhamad$^{\diamondsuit,\spadesuit}$ \quad Yunita Sari$^{\spadesuit}$ \\ $^\diamondsuit$Mantera Studio \quad $^\spadesuit$Universitas Gadjah Mada \\ \texttt{wicaksonoleksonomuhamad2001@mail.ugm.ac.id} \quad \texttt{yunita.sari@ugm.ac.id} \\ \href{https://github.com/airlanggawicaksono/acl-future/}{\faGithub\ Code} }

\begin{document}
\maketitle
\begin{abstract} 
 Classifying implicit hate speech remains a challenge, intent is often masked through insinuation and context rather than explicit slurs. Prior supervised contrastive approaches improve in-domain detection but can overfit surface cues and struggle to transfer across datasets. We propose \textsc{ImpSH}, a triplet-based framework that aligns posts with implied statements when available and uses context-bounded semi-hard negatives to focus learning on near confusions. We also examine \textsc{AugSH}, which forms positives via data augmentation. In controlled evaluations on \textsc{IHC}, \textsc{SBIC}, and \textsc{DynaHate} with \textsc{BERT} and \textsc{HateBERT}, \textsc{ImpSH} is a viable alternative to standard supervised contrastive baselines and often improves cross-domain performance under matched preprocessing and tuning budgets. Representation analysis using alignment and uniformity indicates tighter positive pairs with balanced global spread, and qualitative nearest-neighbor case studies illustrate typical false negatives under domain shift. These results demonstrate that aligning posts with their implied statements via context-bounded mining provides a more stable, bijective-like mapping to related insinuations, overcoming the volatility inherent in traditional clustering-based representation learning.
\end{abstract}

\section*{Content Warning}
The content of this paper may contain offensive, harmful, or distressing language, including examples of hate speech and discriminatory expressions. These materials are included solely for research purposes and do not reflect the views of the authors. Reader discretion is advised.

\section{Introduction}

\begin{figure}[t]
 \centering
 \includegraphics[width=1\linewidth]{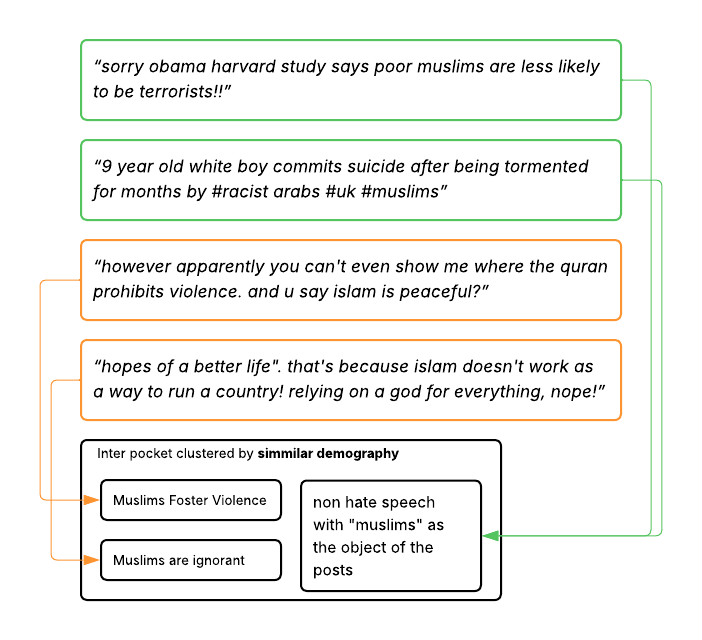}
 \caption{Posts that share similar implied targets form tight yet fragmented clusters; small wording changes disperse points even within the same demographic group \citep{elsherief2021latent}.}
 \label{fig:alignment}
\end{figure}
Classifying implicit hate remains challenging because hateful intent is often expressed indirectly through sarcasm, euphemisms, rhetorical questions, and other pragmatic cues, so surface text features are often insufficient \citep{elsherief2021latent,sap2020social,kim2022generalizable,zhang2024don}. A second challenge is semantic overlap with non-hateful content: posts that target the same group can share topical content despite different labels, which blurs the decision boundary and weakens supervision \citep{sap2020social,elsherief2021latent}. When this boundary is weak, standard training can learn dataset-specific shortcuts tied to topic and group-linked surface cues, reducing cross-dataset generalization and transfer to related abusive-language settings \citep{nejadgholi2020cross,rottger2020hatecheck,sap2019risk}.

Figure~\ref{fig:alignment} illustrates this issue for examples aligned to the same target demographic. Despite different labels, many instances share topic and group mentions and can lie close in representation space, so accurate detection requires inferring intent and implied meaning, not just matching surface wording. To address semantic overlap, prior work uses supervised contrastive learning (SCL), most notably \textsc{ImpCon}. \textsc{ImpCon} pulls each post toward its human-annotated implied statement as a positive pair \citep{khosla2020supervised,gunel2021supervised,kim2022generalizable}. However, standard SCL treats most other in-batch examples as negatives; in implicit hate datasets, many near-neighbors differ only slightly, so repelling them can introduce false negatives and hurt generalization \citep{huynh2020boosting,kalantidis2020hard,wang2019multi}. \citet{ahn2024sharedcon} propose \textsc{SharedCon}, which forms positives from shared semantics among same-label posts, reducing reliance on implied-statement annotations \citep{ahn2024sharedcon}, yet it still inherits the same fragile negative treatment when many examples are topically similar.

Motivated by this remaining fragility on the negative side, we instead target \emph{negative selection}. Through the lens of alignment and uniformity \citep{wang2020understanding}, SCL-style training that aligns positives while repelling all other in-batch instances as negatives \citep{kim2022generalizable,ahn2024sharedcon} can yield tight, topic-driven clusters and weaker global coverage of the embedding space, which can hurt transfer. This effect is amplified by false negatives and class collisions that distort local neighborhoods \citep{chuang2020debiased}. To mitigate clustering driven purely by shared topics and target mentions, we explicitly separate semantically close instances with opposing labels, using a margin-based triplet objective with semi-hard negative mining \citep{schroff2015facenet,hermans2017defense,wu2017sampling,musgrave2020reality,xuan2020improved,robinson2021hardneg}.

In summary, we make the following contributions:
\begin{itemize}
  \item We propose a triplet-based framework for implicit hate detection that uses \emph{context-bounded semi-hard negative mining} to avoid repelling all in-batch negatives, while keeping a standard cross-entropy objective for classification.
  \item We introduce two variants, \textsc{ImpSH} and \textsc{AugSH}. \textsc{ImpSH} uses post-implied-statement positives when available, while \textsc{AugSH} uses augmentation-based positives for all instances to isolate the role of implication.
  \item We evaluate on \textsc{IHC}, \textsc{SBIC}, and \textsc{DynaHate} with \textsc{BERT} and \textsc{HateBERT} under matched tokenization and tuning budgets, and analyze representation structure with alignment and uniformity scores alongside qualitative neighbor and embedding visualizations.
\end{itemize}

\section{Related Work}
Early hate speech detection relied on lexical cues \citep{waseem2016hateful,davidson2017automated}, which often fail for implicit hate expressed via sarcasm, euphemisms, and other pragmatic cues \citep{badjatiya2017deep,golbeck2017large}. More recent benchmarks such as \textsc{IHC} and \textsc{SBIC}, and cross-domain evaluation settings like \textsc{DynaHate}, shifted attention to semantic modeling, but robust cross-dataset generalization remains difficult \citep{sap2020social,elsherief2021latent,vidgen2020learning,kim2022generalizable,ramponi2022features}.

To improve generalization, recent work applies supervised contrastive learning (SCL) to implicit hate. \citet{kim2022generalizable} propose \textsc{ImpCon}, which uses the human-annotated implied statement as a positive and pulls post-implication pairs together. \citet{ahn2024sharedcon} propose \textsc{SharedCon}, which removes reliance on implied statements by mining shared semantics among same-label instances (e.g., clustering) and constructing positives from label-consistent neighborhoods.

\begin{figure}[!t]
 \centering
 \includegraphics[width=1\linewidth]{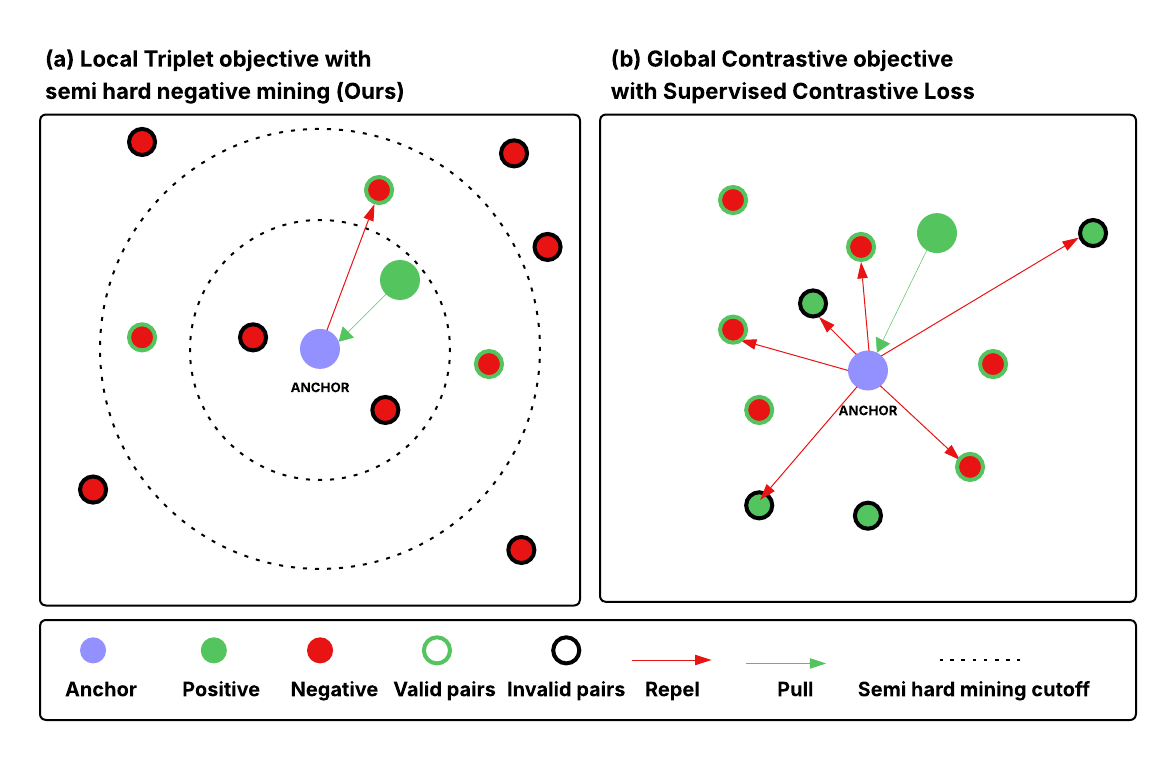}
 \caption{Triplet vs.\ SCL. Triplet updates use a margin-violating, confusable negative; SCL repels all non-positives in the batch.}
 \label{fig:objective}
\end{figure}

\subsection{Triplet Objectives and Negative Selection}
Triplet losses optimize relative comparisons by pulling an anchor toward a positive and pushing it away from a negative by a margin \citep{schroff2015facenet,hermans2017defense}. In contrast, SCL pulls each anchor toward all same-label examples in the minibatch and pushes it away from different-label examples \citep{khosla2020supervised,gunel2021supervised,liao2021sentence}. In datasets with heavy topical overlap, treating most in-batch items as negatives can create false negatives and distort local neighborhoods, which can hurt transfer \citep{huynh2020boosting,kalantidis2020hard,wang2019multi,wang2020understanding}.

Negative selection is therefore critical. Semi-hard mining focuses on negatives that violate the margin but are not extreme outliers, concentrating updates on realistic near-misses \citep{schroff2015facenet,wu2017sampling,musgrave2020reality,xuan2020improved,robinson2021hardneg}. For implicit hate, where examples can share topic and target cues while differing in implied intent, context-bounded semi-hard selection helps the model separate intent-sensitive cases without over-repelling broadly similar content.

\section{Methodology}
\label{sec:methodology}
\begin{figure*}[t]
 \centering
 \includegraphics[width=\textwidth]{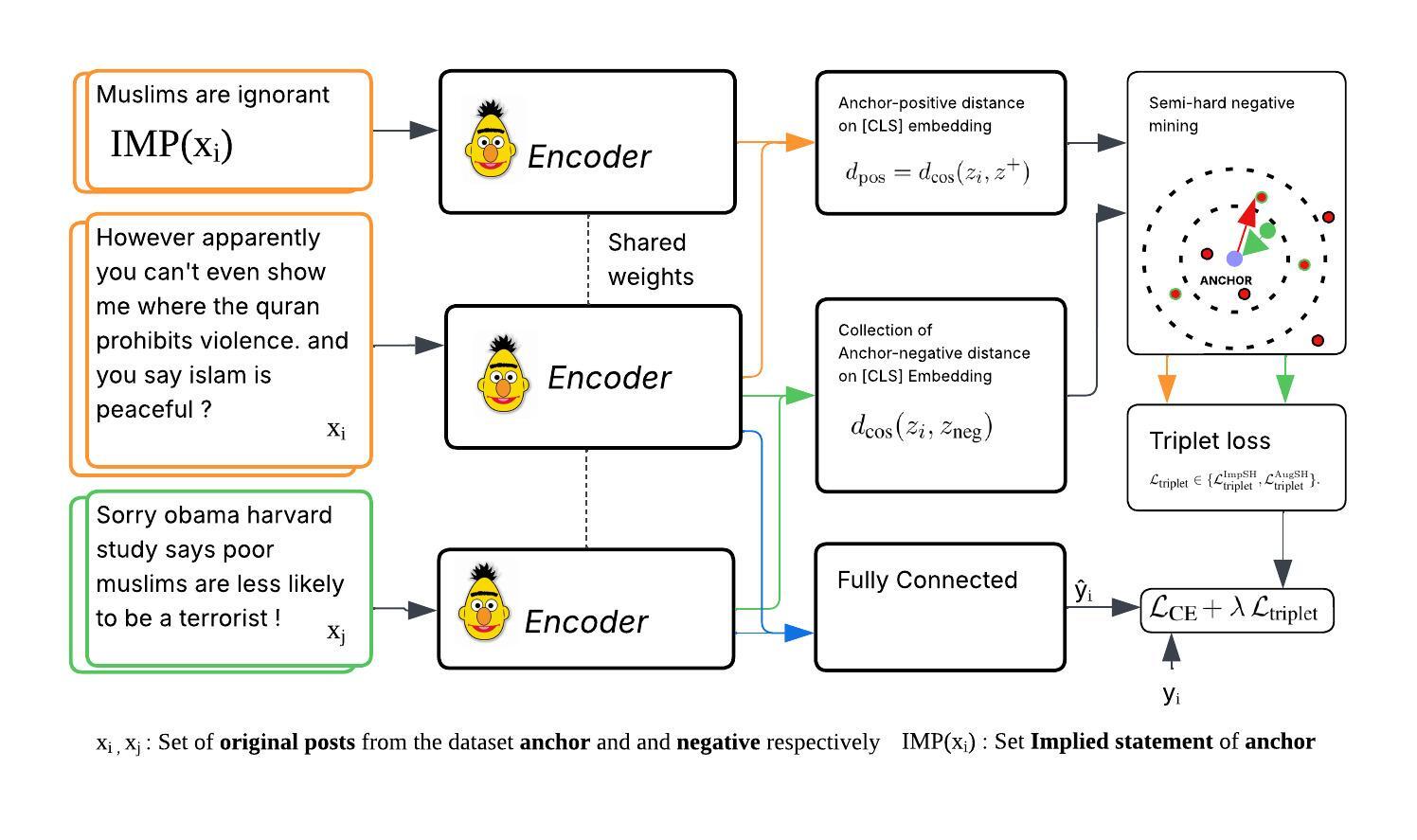}
\caption{Training overview for our triplet framework. Left: positive formation (\textsc{ImpSH} uses a post–implication pair; \textsc{AugSH} uses augmentation-based positives when implication is absent). Right: semi-hard negative mining from the minibatch relative to the chosen anchor–positive pair. The encoder is updated by the triplet loss $\mathcal{L}^{\mathrm{ImpSH}}_{\text{triplet}}$ or $\mathcal{L}^{\mathrm{AugSH}}_{\text{triplet}}$ (Eqs.~\ref{eq:impsh} and~\ref{eq:augsh}), alongside a standard classification loss $\mathcal{L}_{\mathrm{CE}}$.}
\label{fig:overview}

\end{figure*}
Figure~\ref{fig:overview} summarizes training. We jointly optimize a standard cross-entropy loss
$\mathcal{L}_{\mathrm{CE}}$ and a triplet loss, mining semi-hard negatives within each minibatch
(Eqs.~\ref{eq:impsh}-\ref{eq:augsh}). For each anchor-positive pair, we select an opposite-label
negative that is farther than the positive but still within the margin band, focusing updates on
near-confusable cases.

The standard cross-entropy loss is:
\begin{equation}
\label{eq:ce}
\mathcal{L}_{\text{CE}} = - \frac{1}{N} \sum_{i=1}^N \left[ y_i \log \hat{y}_i + (1 - y_i) \log (1 - \hat{y}_i) \right].
\end{equation}

We use cosine distance:
\begin{equation}
\label{eq:cos}
d_{\text{cos}}(z_i, z_j) = 1 - \frac{z_i \cdot z_j}{\lVert z_i \rVert \, \lVert z_j \rVert}.
\end{equation}

We define one positive per anchor. For hateful $x_i$, $z_i^{+}=h(\mathrm{IMP}(x_i))$; for non-hate $x_i$,
$z_i^{+}=h(\mathrm{AUG}(x_i))$, where $h(\cdot)$ is the encoder, $\mathrm{IMP}(\cdot)$ returns the implied
statement, and $\mathrm{AUG}(\cdot)$ applies synonym augmentation. The anchor-positive distance is:
\begin{equation}
\label{eq:dpos}
d_{\mathrm{pos}} = d_{\text{cos}}(z_i, z_i^{+}).
\end{equation}

To ensure the model separates negatives that are truly close to the positive, we select a context-bounded semi-hard negative from the opposite-label pool $\mathcal{N}$:
\begin{equation}
\label{eq:neg}
z_{\text{neg}} = \underset{
 \substack{z_j \in \mathcal{N} \\ d_{\mathrm{pos}} < d_{\text{cos}}(z_i, z_j) < d_{\mathrm{pos}} + \alpha}
}{\arg\min}\; d_{\text{cos}}(z_i, z_j).
\end{equation}

The triplet objectives specialize by the choice of $z^{+}$:
\begin{equation}
\begin{aligned}
\mathcal{L}^{\mathrm{ImpSH}}_{\text{triplet}}
&= \frac{1}{N}\sum_{i=1}^{N}\mathrm{ReLU}\!\left[ d_{\mathrm{pos}} - d_{\text{cos}}(z_i, z_{\text{neg}}) + \alpha \right] \\
&\text{with } z^{+}=h(\mathrm{IMP}(x_i))
\end{aligned}
\label{eq:impsh}
\end{equation}

\begin{equation}
\begin{aligned}
\mathcal{L}^{\mathrm{AugSH}}_{\text{triplet}}
&= \frac{1}{N}\sum_{i=1}^{N}\mathrm{ReLU}\!\left[ d_{\mathrm{pos}} - d_{\text{cos}}(z_i, z_{\text{neg}}) + \alpha \right] \\
&\text{with } z^{+}=h(\mathrm{AUG}(x_i))
\end{aligned}
\label{eq:augsh}
\end{equation}

The final training objective is:
\begin{equation}
\label{eq:final_loss}
\mathcal{L} = \mathcal{L}_{\text{CE}} + \lambda\, \mathcal{L}_{\text{triplet}},
\mathcal{L}_{\text{triplet}} \in \{\mathcal{L}^{\mathrm{ImpSH}}_{\text{triplet}},
                  \mathcal{L}^{\mathrm{AugSH}}_{\text{triplet}}\}.
\end{equation}
For ImpSH \ref{eq:impsh} method we use implied statement if available, otherwise we use augmentation. And for AugSH we use augmentation for both label.

\section{Experiments}
\label{sec:experiment}
\subsection{Dataset}
Following \citet{kim2022generalizable,ahn2024sharedcon}, we treat implicit hate detection as a binary classification task. We evaluate on three datasets shown in Table~\ref{tab:dataset}. \textsc{IHC} \citep{elsherief2021latent} contains tweets labeled for hate, with annotations for target groups and implied hateful meanings. \textsc{SBIC} \citep{sap2020social} contains Reddit posts annotated for social bias and offensiveness, and it also provides free-text implications. \textsc{DynaHate} \citep{vidgen2020learning} is a dynamically generated hate speech dataset created through human-model-in-the-loop collection, with diverse and challenging hate examples. For \textsc{SBIC} and \textsc{DynaHate}, we map all abusive categories (explicit/implicit/offensive) to the \emph{hate} label for binary evaluation.

\begin{table}[h!]
\centering
\begin{tabular}{lccc}
\hline
Dataset & Total & Non-hate & Hate \\
\hline
\textsc{IHC} & 18,666 & 13,206 & 5,460 \\
\textsc{SBIC} & 44,875 & 21,276 & 23,599 \\
\textsc{DynaHate} & 33,006 & 17,804 & 15,202 \\
\hline
\end{tabular}
\caption{Datasets used for training and cross-dataset evaluation, including class distribution. The \textit{hate} label covers offensive, explicit, and implicit hate.}
\label{tab:dataset}
\end{table}

\subsection{Implementation Details}
\label{sec:implement}
We fine-tune BERT~\citep{devlin2019bert} and HateBERT~\citep{caselli2021hatebert} as sentence encoders. Following \citet{kim2022generalizable}, we fix the learning-rate-to-batch-size ratio, using a batch size of 8 and a learning rate of $2 \times 10^{-5}$ with AdamW. Models are trained for 6 epochs with dropout 0.1 on RTX 3050 GPU (4GB). For the metric-learning objective, we tune $\lambda = 0.25$ and $\alpha \in \{0.1, 0.2, 0.3, 0.4, 0.5\}$ using dev set performance. Experiments use 4 random seeds $\{0, 1, 2, 3\}$. Synonym substitution is applied via NLPAug\footnote{\url{https://nlpaug.readthedocs.io/en/latest/augmenter/word/synonym.html}} as in \citet{kim2022generalizable}.

\subsection{Baselines}
We compare against four objectives:
\begin{itemize}
\item \textbf{CE.} Standard fine-tuning with cross-entropy only.
\item \textbf{CE + SCL} \citep{gunel2021supervised}. Cross-entropy plus supervised contrastive loss.
\item \textbf{CE + ImpCon} \citep{kim2022generalizable}. Cross-entropy plus implication-aware supervised contrastive loss.
\item \textbf{CE + SharedCon} \citep{ahn2024sharedcon}. Cross-entropy plus shared-semantics supervised contrastive loss.
\end{itemize}
All baselines use the same backbone, preprocessing, and optimization setup as our method.

\section{Results and Analysis}
\label{sec:results}
\subsection{F1 Score result }
\begin{table*}[!ht]
\centering
\begin{tabular}{lllll}
\hline
\textbf{Model} & \textbf{Objective} & \textsc{IHC}$\rightarrow$\textsc{SBIC} & \textsc{IHC}$\rightarrow$\textsc{DynaHate} & \textsc{IHC}$\rightarrow$\textsc{IHC} \\
\hline
BERT & CE        & 56.9\% & 53.3\% & 77.7\% \\
BERT & CE+SCL~\citep{gunel2021supervised}  & 59.8\% & 52.2\% & 77.7\% \\
BERT & CE+ImpCon~\citep{kim2022generalizable} & 60.8\% & 57.9\% & 78.0\% \\
BERT & CE+SharedCon~\citep{ahn2024sharedcon} & \textbf{65.2}\% & 59.1\% & \textbf{78.4\%} \\

\hline
BERT & CE+AugSH      & 59.2\% & 54.8\% & 77.8\% \\
BERT & CE+ImpSH      & 61.4\% & \textbf{59.2\%} & 78.3\% \\
\hline
HateBERT & CE       & 58.8\% & 54.8\% & 76.5\% \\
HateBERT & CE+SCL~\citep{gunel2021supervised} & 55.9\% & 52.7\% & 76.9\% \\
HateBERT & CE+ImpCon~\citep{kim2022generalizable} & 63.9\% & 59.5\% & \textbf{77.4\%} \\
HateBERT & CE+SharedCon~\citep{ahn2024sharedcon} &63.5\% &57.7\% &77.1\% \\

\hline
HateBERT & CE+AugSH     & 58.2\% & 54.8\% & 77.2\% \\
HateBERT & CE+ImpSH     & \textbf{65.0\%} & \textbf{60.8\%} & 76.4\% \\
\hline
 & & \textsc{SBIC}$\rightarrow$\textsc{IHC} & \textsc{SBIC}$\rightarrow$\textsc{DynaHate} & \textsc{SBIC}$\rightarrow$\textsc{SBIC} \\
\hline
BERT & CE        & 59.7\% & 60.3\% & 83.7\% \\
BERT & CE+SCL~\citep{gunel2021supervised}  & 59.4\% & 60.9\% & 83.7\% \\
BERT & CE+ImpCon~\citep{kim2022generalizable} & 61.4\% & 61.2\% & 83.8\% \\
BERT & CE+SharedCon~\citep{ahn2024sharedcon} &\textbf{62.5\%} &\textbf{62.0\%} &83.8\% \\

\hline
BERT & CE+AugSH      & 58.4\% & 60.6\%& 82.3\% \\
BERT & CE+ImpSH      & 61.8\% & \textbf{62.0\%} & \textbf{84.1\%} \\
\hline
HateBERT & CE       & 59.0\% & 60.2\% & 84.1\% \\
HateBERT & CE+SCL~\citep{gunel2021supervised} & 59.5\% & 59.3\% & 84.3\% \\
HateBERT & CE+ImpCon~\citep{kim2022generalizable} & 60.1\% & 60.4\% & \textbf{84.8\%} \\
HateBERT & CE+SharedCon~\citep{ahn2024sharedcon} & 59.3\% & 59.4\% & 84.5\% \\
\hline
HateBERT & CE+AugSH     & 60.0\% & \textbf{60.6}\% & 84.4\% \\
HateBERT & CE+ImpSH     & \textbf{60.6\%} & 60.4\% & 84.4\% \\
\hline
\end{tabular}
\caption{\label{tab:f1-result}
F1-scores for in-domain and cross-dataset evaluations. Models are trained on \textsc{IHC} or \textsc{SBIC} and tested on the dataset indicated by the arrow. Bold marks the best objective \emph{within each encoder} for a given source$\rightarrow$target; ties are both bolded. See Appendix~\ref{sec:seed-rep} for details.}
\end{table*}
Table~\ref{tab:f1-result} reports macro-F1 (\%) averaged over four seeds described in Section~\ref{sec:implement} with both in-domain and cross-dataset evaluation. Since the goal of \textsc{ImpSH} is to handle semantic overlap during representation learning for generalization, we compare \textsc{ImpSH} primarily on cross-dataset transfer and we use \textsc{AugSH} as an ablation that keeps augmentation based multi-view training but removes implied-statement supervision. Prior work finds that in-domain and out-of-distribution performance can be strongly correlated but can also become inversely related on real benchmarks, so selecting models only by in-domain performance can miss the best cross-dataset model \citep{miller2021accuracy,teney2023id}. With BERT trained on \textsc{IHC}, \textsc{ImpSH} is best for transfer to \textsc{DynaHate} but it does not surpass the strongest baseline for transfer to \textsc{SBIC} and it also does not surpass the strongest baseline for in-domain evaluation on \textsc{IHC}, where \textsc{SharedCon} remains best. In both cases \textsc{ImpSH} still surpasses the next strongest baseline which is \textsc{ImpCon}, and it remains above \textsc{AugSH} across all three evaluations, which supports that implied-statement supervision adds signal beyond augmentation alone. While \textsc{SharedCon} shows high scores in specific settings, it relies on clustering-based shared semantics which is often volatile and sensitive to batch composition. In contrast, \textsc{ImpSH} maintains more consistent alignment by mining negatives specifically relative to the context of the implied statement, focusing on separating negatives that are semantically close to the positive. With HateBERT trained on \textsc{IHC}, \textsc{ImpSH} is strongest on both transfer targets but it does not surpass the strongest in-domain baseline, where \textsc{ImpCon} is best and \textsc{AugSH} is second, and \textsc{ImpSH} falls below both. This suggests a transfer versus in-domain tension in this configuration, where objectives that best separate the source dataset can differ from objectives that best preserve meaning under cross-dataset shift, while the transfer gap between \textsc{ImpSH} and \textsc{AugSH} still supports benefits beyond augmentation on both targets. 

With BERT trained on \textsc{SBIC}, \textsc{ImpSH} does not surpass the strongest baseline for transfer to \textsc{IHC}, where \textsc{SharedCon} remains best, but it surpasses the next strongest baseline which is \textsc{ImpCon} and it also stays above \textsc{AugSH}. For transfer to \textsc{DynaHate}, \textsc{ImpSH} matches the strongest baseline which is \textsc{SharedCon} and it stays above the next strongest baseline which is \textsc{ImpCon}. For in-domain evaluation on \textsc{SBIC}, \textsc{ImpSH} is the strongest objective and it surpasses the next strongest baselines, \textsc{ImpCon} and \textsc{SharedCon}, and it remains above \textsc{AugSH}, which again supports that implied-statement supervision contributes beyond augmentation. 

With HateBERT trained on \textsc{SBIC}, \textsc{ImpSH} is best for transfer to \textsc{IHC} and it surpasses the next strongest baseline which is \textsc{ImpCon} while remaining above \textsc{AugSH}. For transfer to \textsc{DynaHate}, \textsc{ImpSH} does not surpass the strongest baseline, where \textsc{AugSH} is best, and \textsc{ImpSH} matches the next strongest baseline which is \textsc{ImpCon}. This is the one setting where the ablation suggests that multi-view regularization accounts for most of the gain and implied-statement supervision adds limited additional benefit. For in-domain evaluation on \textsc{SBIC} with HateBERT, \textsc{ImpSH} does not surpass the strongest baseline, where \textsc{ImpCon} is best, and it also does not surpass the next strongest baseline, where \textsc{SharedCon} is second, while it matches \textsc{AugSH}, which is consistent with implied-statement supervision being less helpful for maximizing in-domain fit on this source.

\subsection{Alignment and uniformity}
Following standard practice for representation evaluation~\citep{wang2020understanding}, we assess \textbf{Alignment} and \textbf{Uniformity} on our best encoder, \textsc{HateBERT}~\citep{caselli2021hatebert}. We L2-normalize embeddings so that $\|f(x)\|_2=1$. Alignment measures how close a positive pair lands in the embedding space, and Uniformity measures how evenly the full set of normalized embeddings spreads on the hypersphere, which helps detect representation collapse. Lower is better for both metrics, and for Uniformity this typically appears as more negative values~\citep{wang2020understanding}. We use $r=2$ for Alignment and report a global score by averaging per-class values, and we use $t=2$ for Uniformity.

This analysis supports our hypothesis because improved cross-domain generalization should appear as tighter positive neighborhoods without sacrificing global spread. Table~\ref{tab:align} shows that \textsc{ImpSH} achieves the best Alignment in four of six train$\rightarrow$test settings, specifically \textsc{IHC}$\rightarrow$\textsc{IHC}, \textsc{IHC}$\rightarrow$\textsc{SBIC}, \textsc{SBIC}$\rightarrow$\textsc{SBIC}, and \textsc{SBIC}$\rightarrow$\textsc{IHC}. The remaining two cases are \textsc{DynaHate} evaluations, where \textsc{SharedCon} attains lower Alignment, which is consistent with its clustering-based objective that pulls together shared semantics and can be effective under perturbation-heavy shifts~\citep{ahn2024sharedcon,vidgen2020learning}. For Uniformity, \textsc{ImpSH} is best in three of six settings, including both transfers to \textsc{DynaHate}, while \textsc{ImpCon} remains strongest on several in-domain or near-domain evaluations. Overall, \textsc{ImpSH} tends to improve local compactness under transfer while maintaining a competitive global spread, and this trend is qualitatively consistent with our t\textsc{-}SNE inspection.

\begin{table}[t]
\centering
\small
\setlength{\tabcolsep}{3pt}
\renewcommand{\arraystretch}{1.05}
\resizebox{\columnwidth}{!}{%
\begin{tabular}{@{}ll|ccc@{}}
\toprule
& \textbf{Dataset} & \textbf{ImpCon} & \textbf{SharedCon} & \textbf{ImpSH (Ours)} \\
\midrule
\multirow{8}{*}{\rotatebox{90}{\textbf{Align.} $\downarrow$}}
& \multicolumn{4}{c}{\textit{Trained on \textsc{IHC}}} \\
& \textsc{IHC}      & 1.925 & 1.785 & \textbf{1.741} \\
& \textsc{SBIC}     & 1.736 & 1.663 & \textbf{1.659} \\
& \textsc{DynaHate} & 1.683 & \textbf{1.423} & 1.754 \\
\cmidrule(lr){2-5}
& \multicolumn{4}{c}{\textit{Trained on \textsc{SBIC}}} \\
& \textsc{SBIC}     & 1.590 & 1.738 & \textbf{1.457} \\
& \textsc{IHC}      & 1.910 & 1.689 & \textbf{1.386} \\
& \textsc{DynaHate} & 1.870 & \textbf{1.222} & 1.395 \\
\midrule
\multirow{8}{*}{\rotatebox{90}{\textbf{Unif.} $\downarrow$}}
& \multicolumn{4}{c}{\textit{Trained on \textsc{IHC}}} \\
& \textsc{IHC}      & \textbf{-3.471} & -2.695 & -3.373 \\
& \textsc{SBIC}     & \textbf{-3.104} & -2.593 & -2.925 \\
& \textsc{DynaHate} & -2.613 & -2.322 & \textbf{-3.351} \\
\cmidrule(lr){2-5}
& \multicolumn{4}{c}{\textit{Trained on \textsc{SBIC}}} \\
& \textsc{SBIC}     & -2.550 & -2.803 & \textbf{-2.891} \\
& \textsc{IHC}      & \textbf{-3.170} & -2.431 & -2.657 \\
& \textsc{DynaHate} & -2.620 & -1.766 & \textbf{-2.765} \\
\bottomrule
\end{tabular}%
}
\caption{Averaged Alignment and Uniformity scores across 4 seeds comparison across methods trained on HateBERT Encoder. Lower is better ($\downarrow$). Best results in bold.}\label{tab:align}
\end{table}

\subsection{Qualitative Representation Analysis}
\begin{figure*}[!ht]
\centering
\setlength{\tabcolsep}{1.5pt}
\begin{tabular}{@{}c ccc ccc@{}}
& \multicolumn{3}{c}{\small Class View} & \multicolumn{3}{c}{\small Target View} \\[0.2em]
& {\scriptsize ImpCon} & {\scriptsize SharedCon} & {\scriptsize ImpSH} 
& {\scriptsize ImpCon} & {\scriptsize SharedCon} & {\scriptsize ImpSH} \\[0.5em]

\rotatebox{90}{\scriptsize\textsc{IHC}\hspace{1em}} &
\begin{subfigure}[b]{0.14\textwidth}\centering\includegraphics[width=\linewidth]{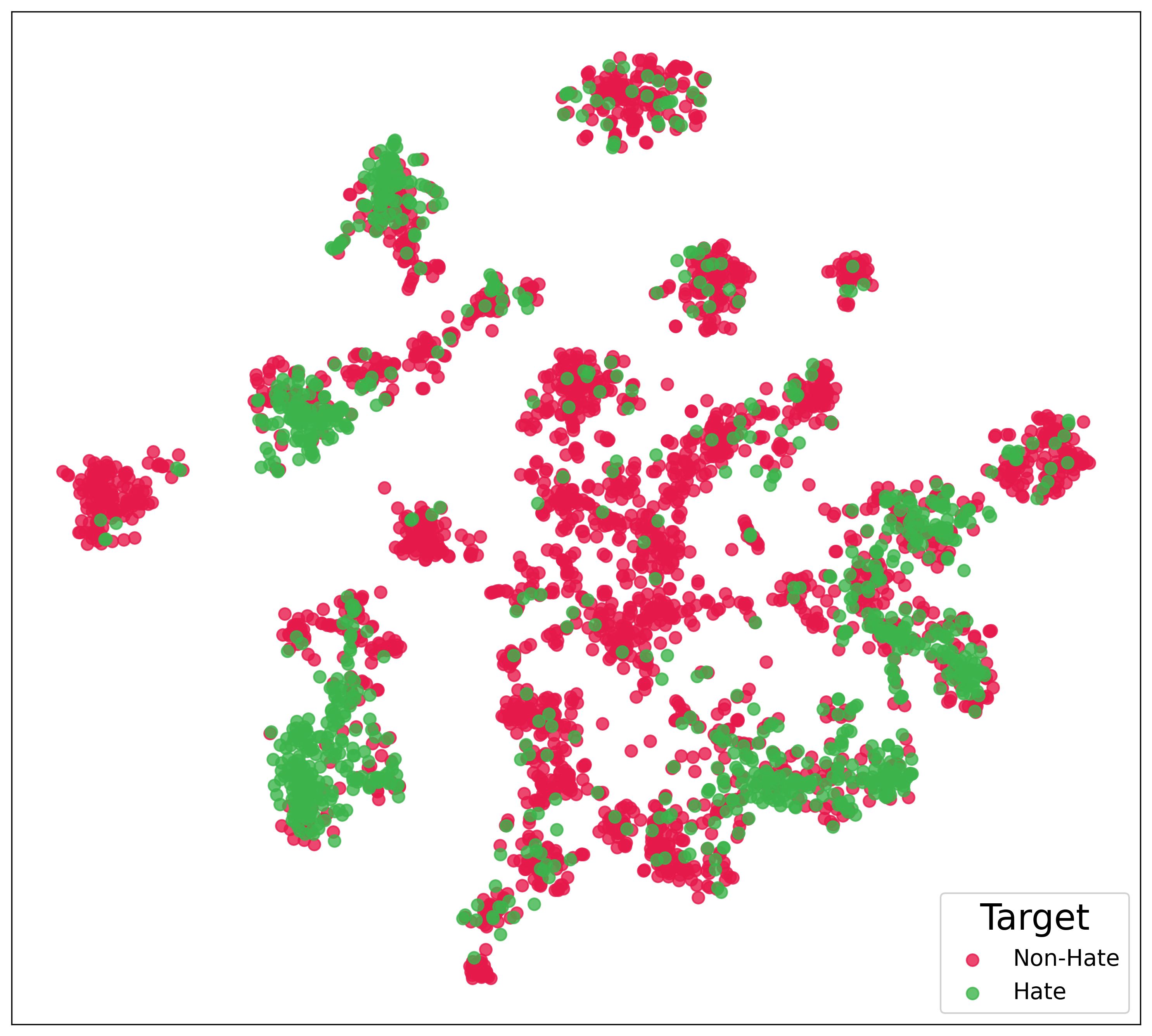}\caption{}\label{fig:qual_impcon_ihc_class}\end{subfigure} &
\begin{subfigure}[b]{0.14\textwidth}\centering\includegraphics[width=\linewidth]{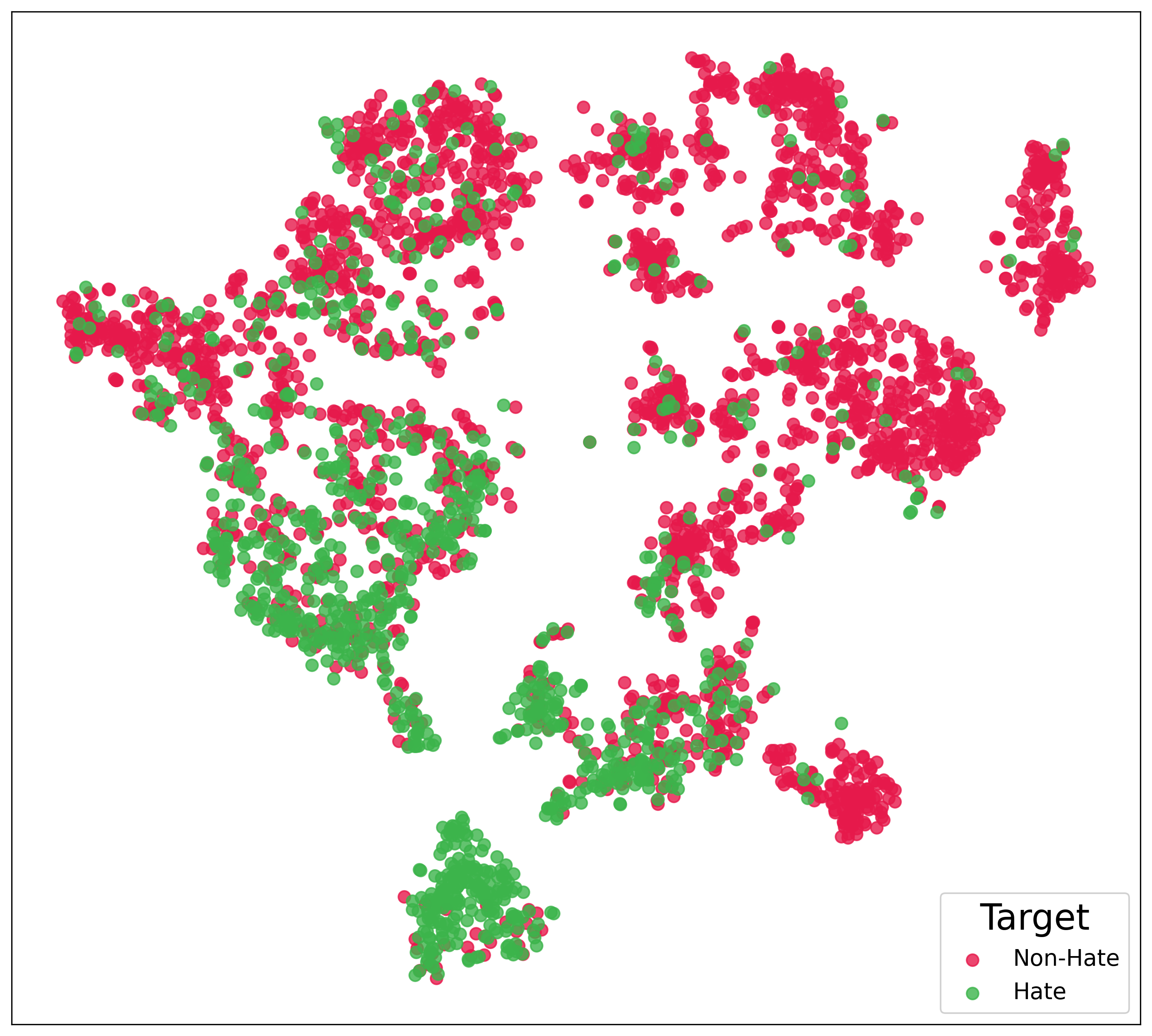}\caption{}\label{fig:qual_shared_ihc_class}\end{subfigure} &
\begin{subfigure}[b]{0.14\textwidth}\centering\includegraphics[width=\linewidth]{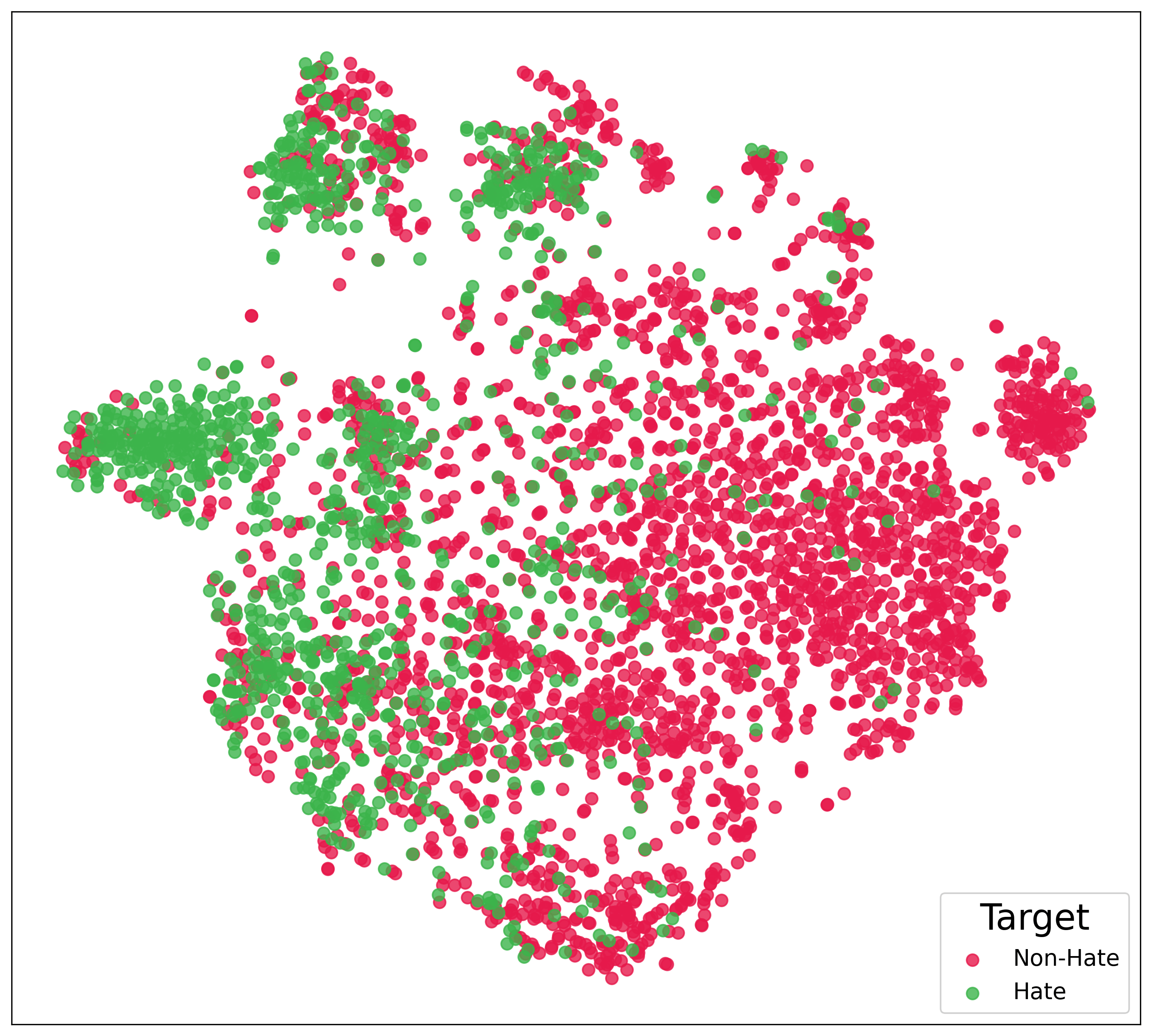}\caption{}\label{fig:qual_impsh_ihc_class}\end{subfigure} &
\begin{subfigure}[b]{0.14\textwidth}\centering\includegraphics[width=\linewidth]{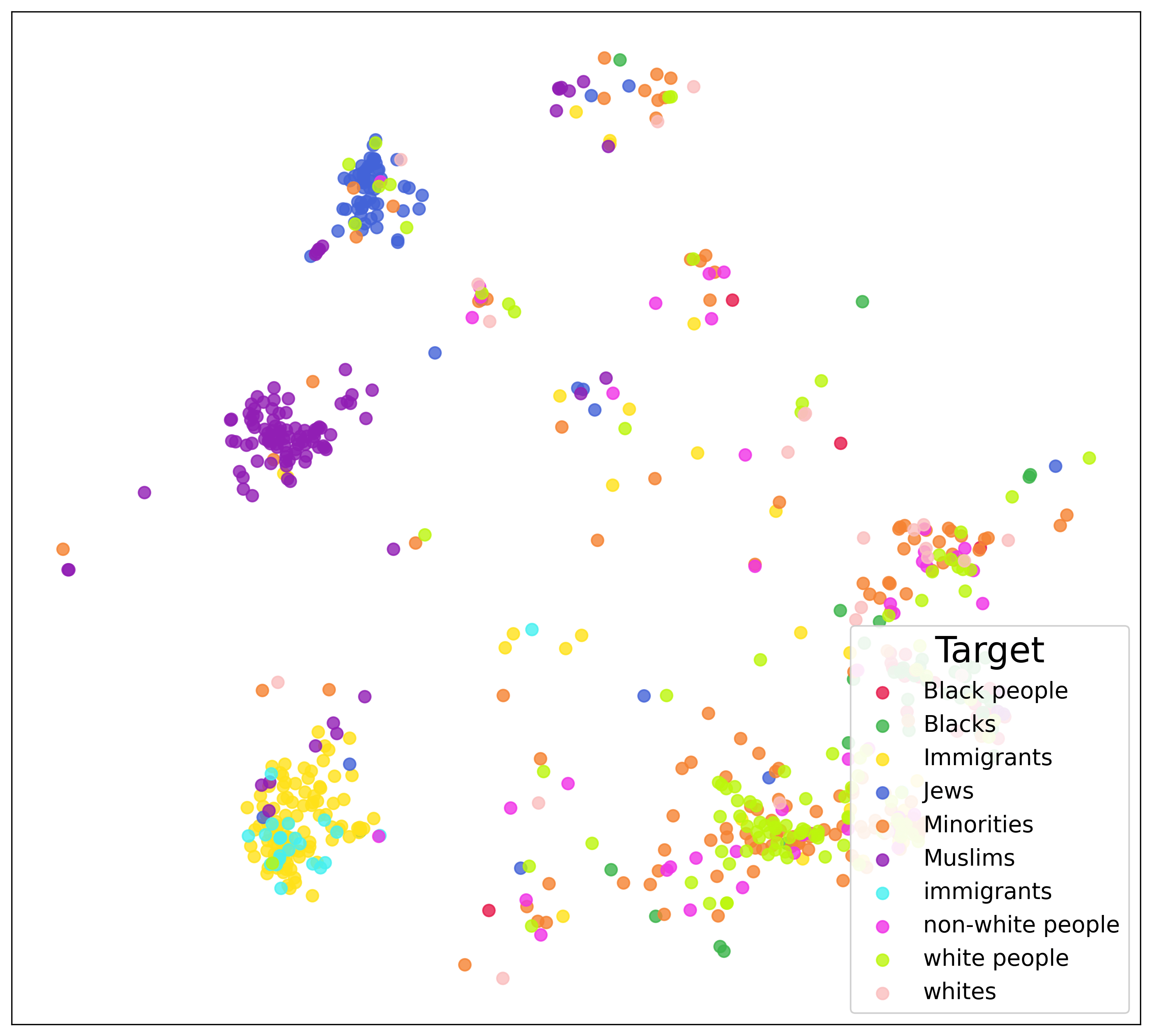}\caption{}\label{fig:qual_impcon_ihc_target}\end{subfigure} &
\begin{subfigure}[b]{0.14\textwidth}\centering\includegraphics[width=\linewidth]{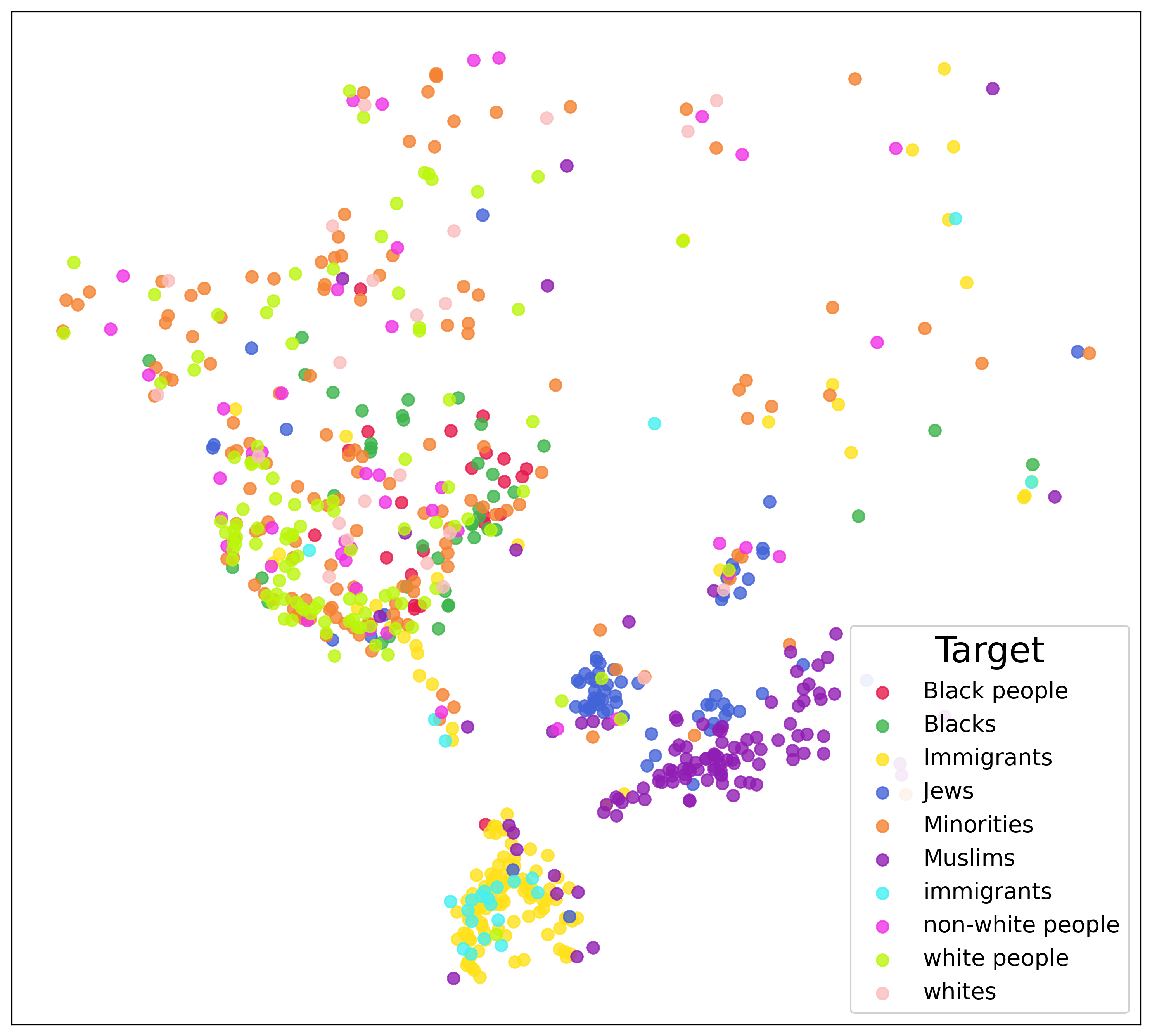}\caption{}\label{fig:qual_shared_ihc_target}\end{subfigure} &
\begin{subfigure}[b]{0.14\textwidth}\centering\includegraphics[width=\linewidth]{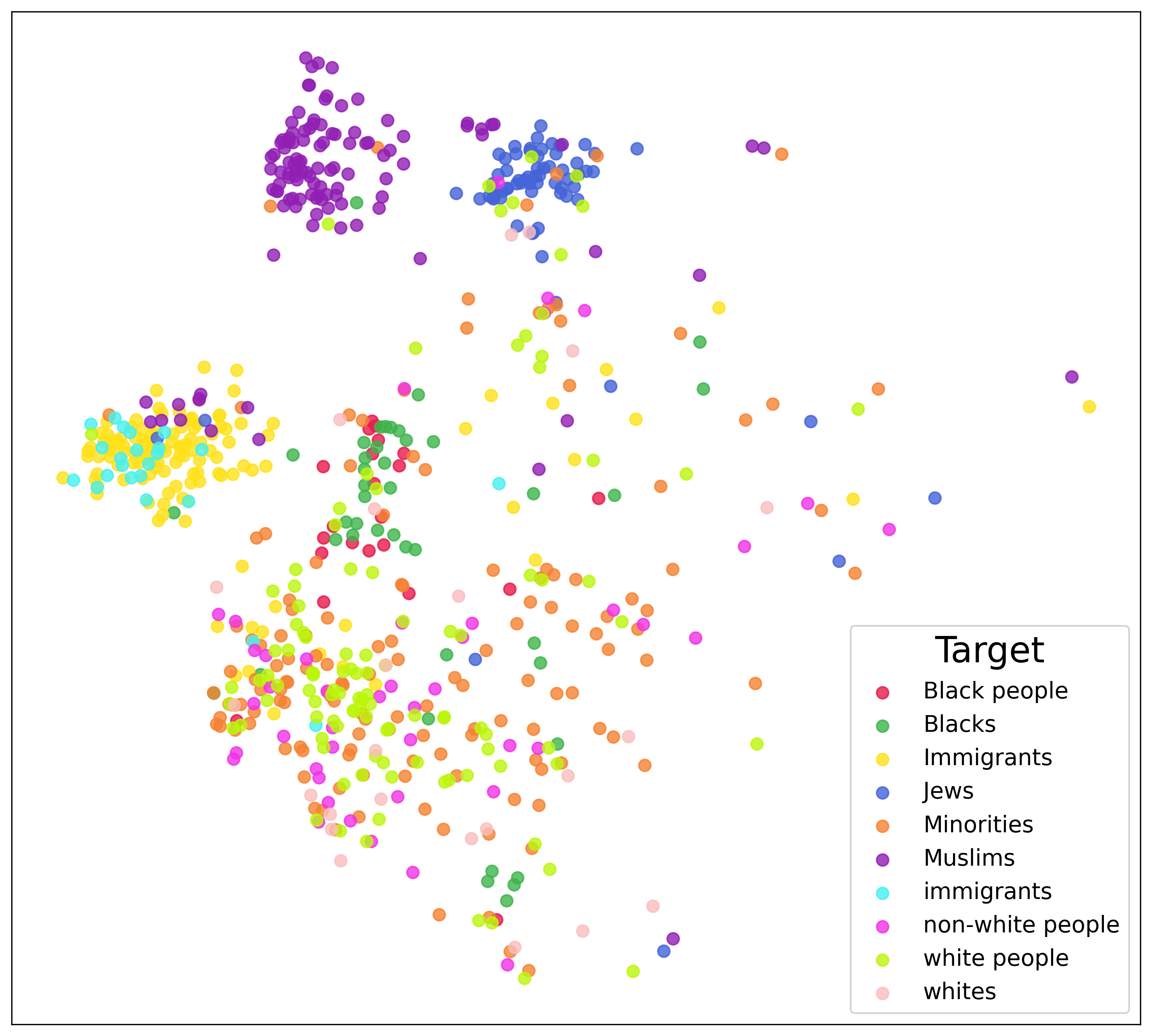}\caption{}\label{fig:qual_impsh_ihc_target}\end{subfigure} \\[0.3em]

\rotatebox{90}{\scriptsize\textsc{SBIC}\hspace{1em}} &
\begin{subfigure}[b]{0.14\textwidth}\centering\includegraphics[width=\linewidth]{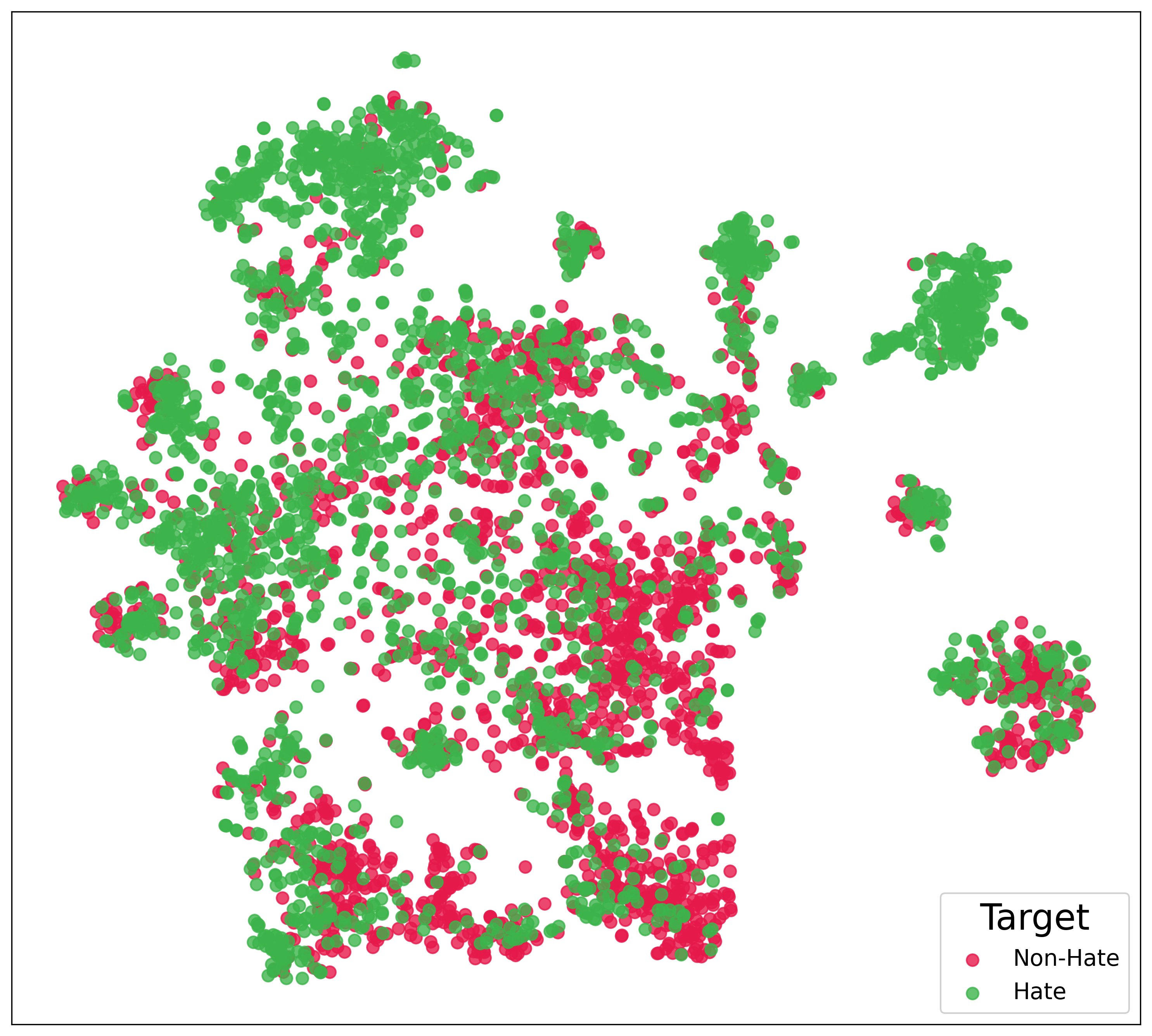}\caption{}\label{fig:qual_impcon_sbic_class}\end{subfigure} &
\begin{subfigure}[b]{0.14\textwidth}\centering\includegraphics[width=\linewidth]{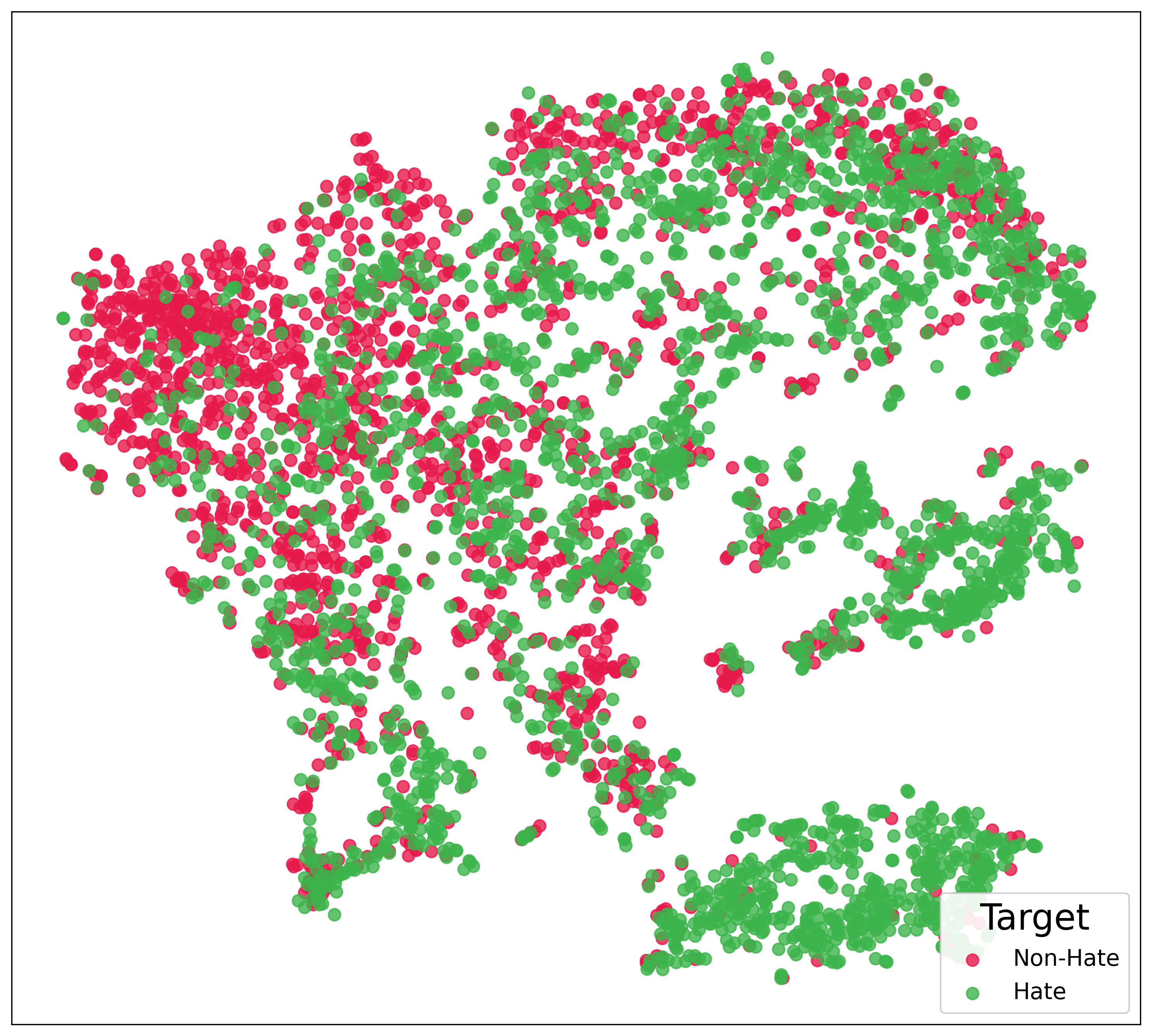}\caption{}\label{fig:qual_shared_sbic_class}\end{subfigure} &
\begin{subfigure}[b]{0.14\textwidth}\centering\includegraphics[width=\linewidth]{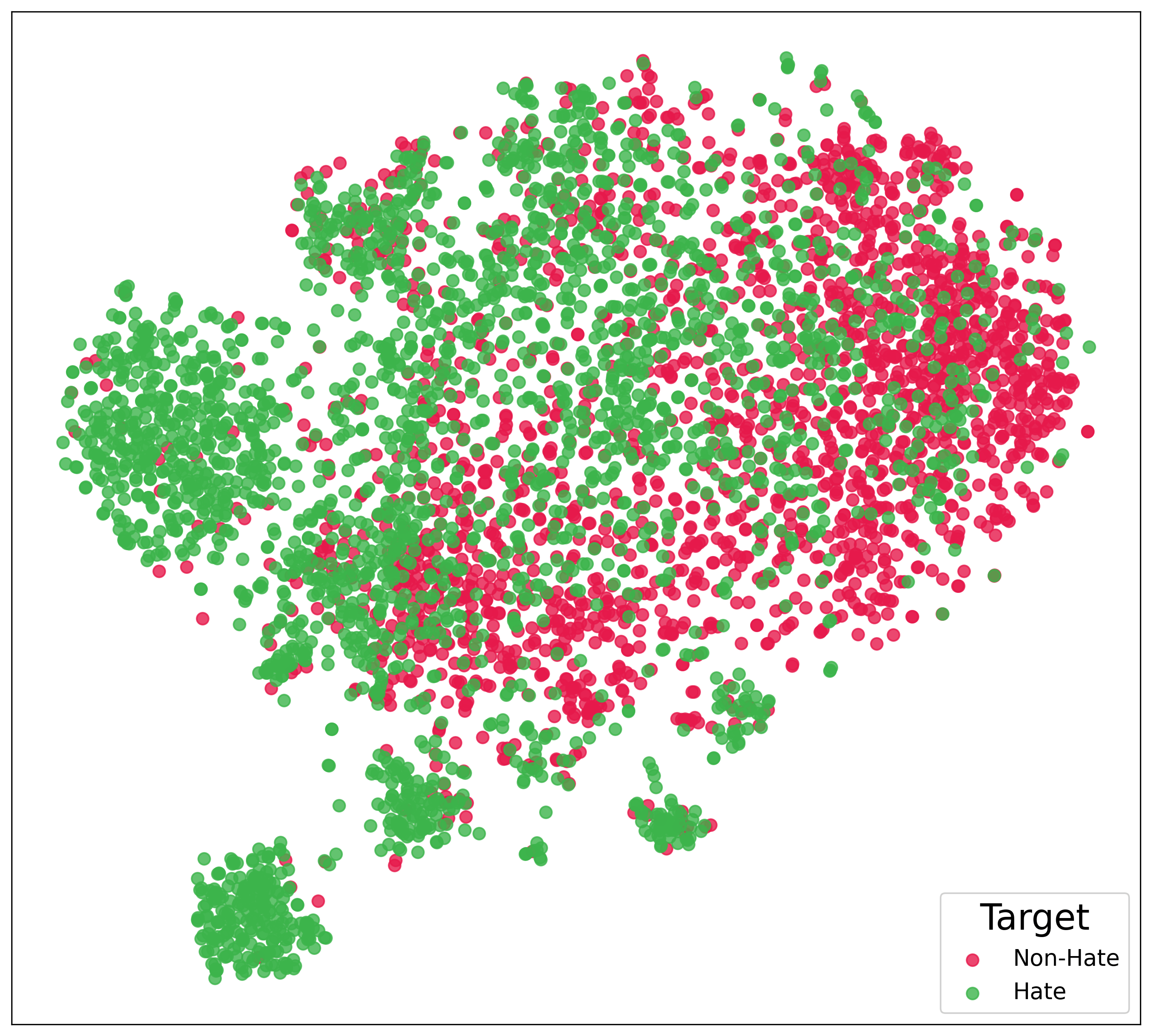}\caption{}\label{fig:qual_impsh_sbic_class}\end{subfigure} &
\begin{subfigure}[b]{0.14\textwidth}\centering\includegraphics[width=\linewidth]{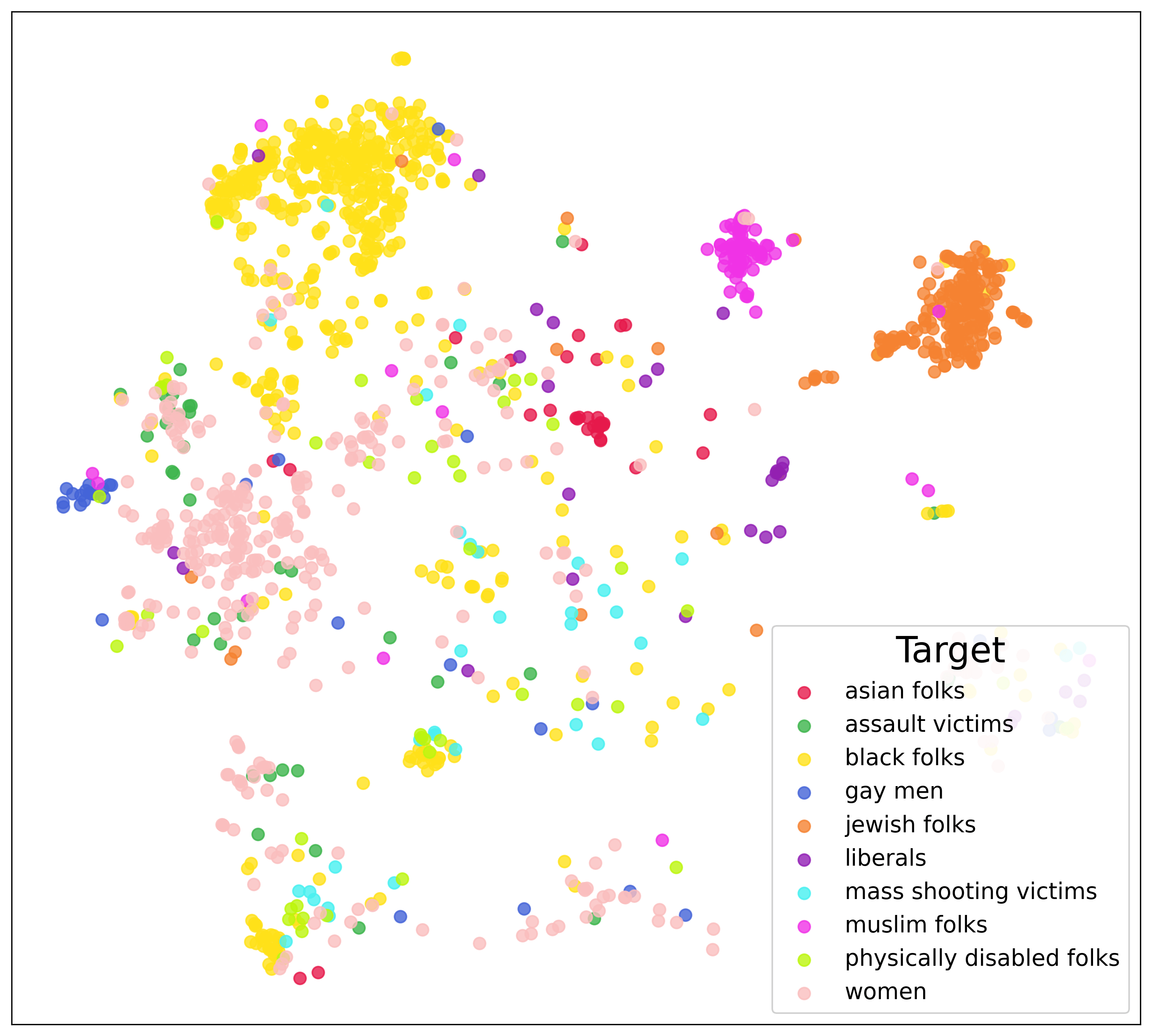}\caption{}\label{fig:qual_impcon_sbic_target}\end{subfigure} &
\begin{subfigure}[b]{0.14\textwidth}\centering\includegraphics[width=\linewidth]{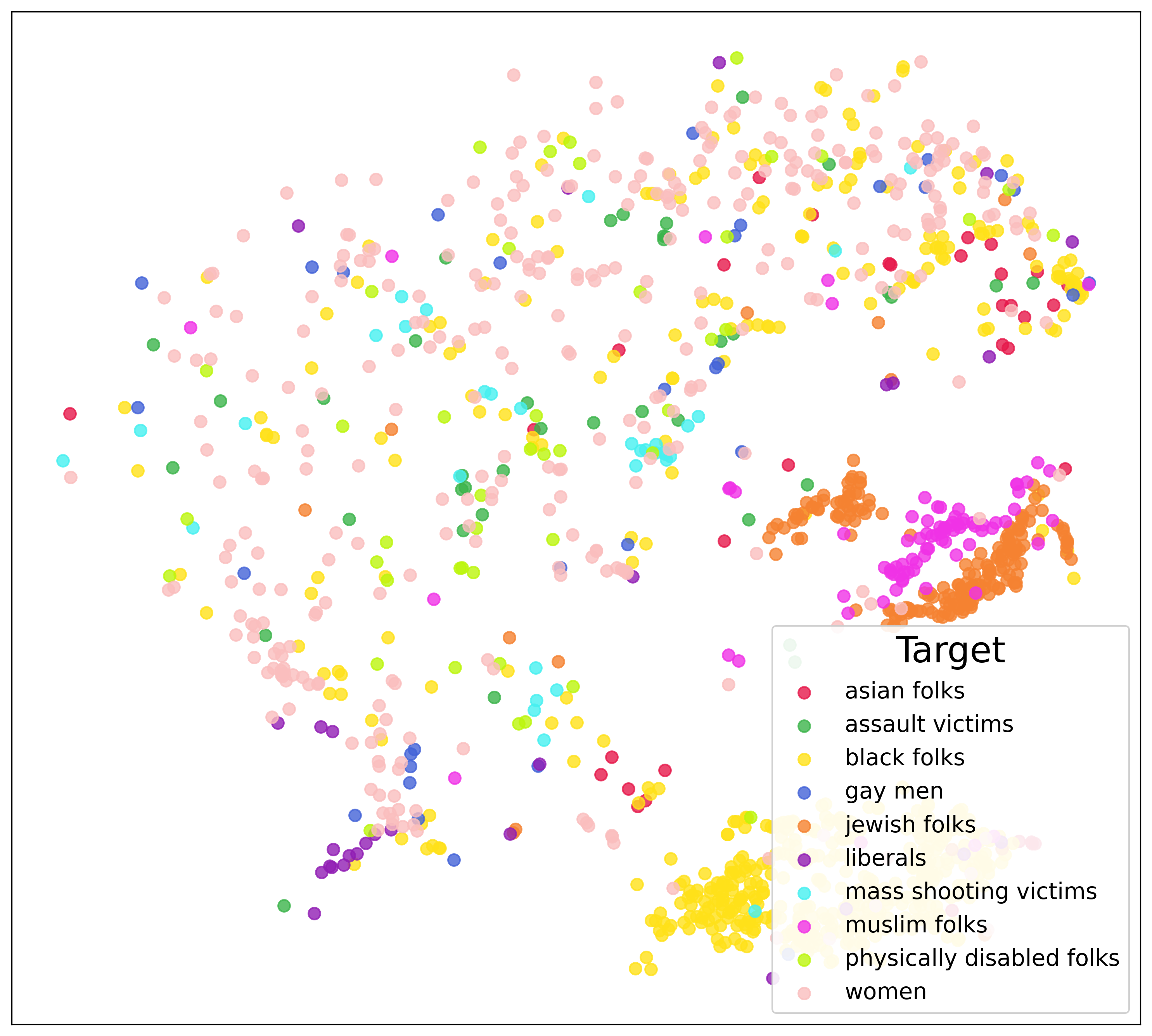}\caption{}\label{fig:qual_shared_sbic_target}\end{subfigure} &
\begin{subfigure}[b]{0.14\textwidth}\centering\includegraphics[width=\linewidth]{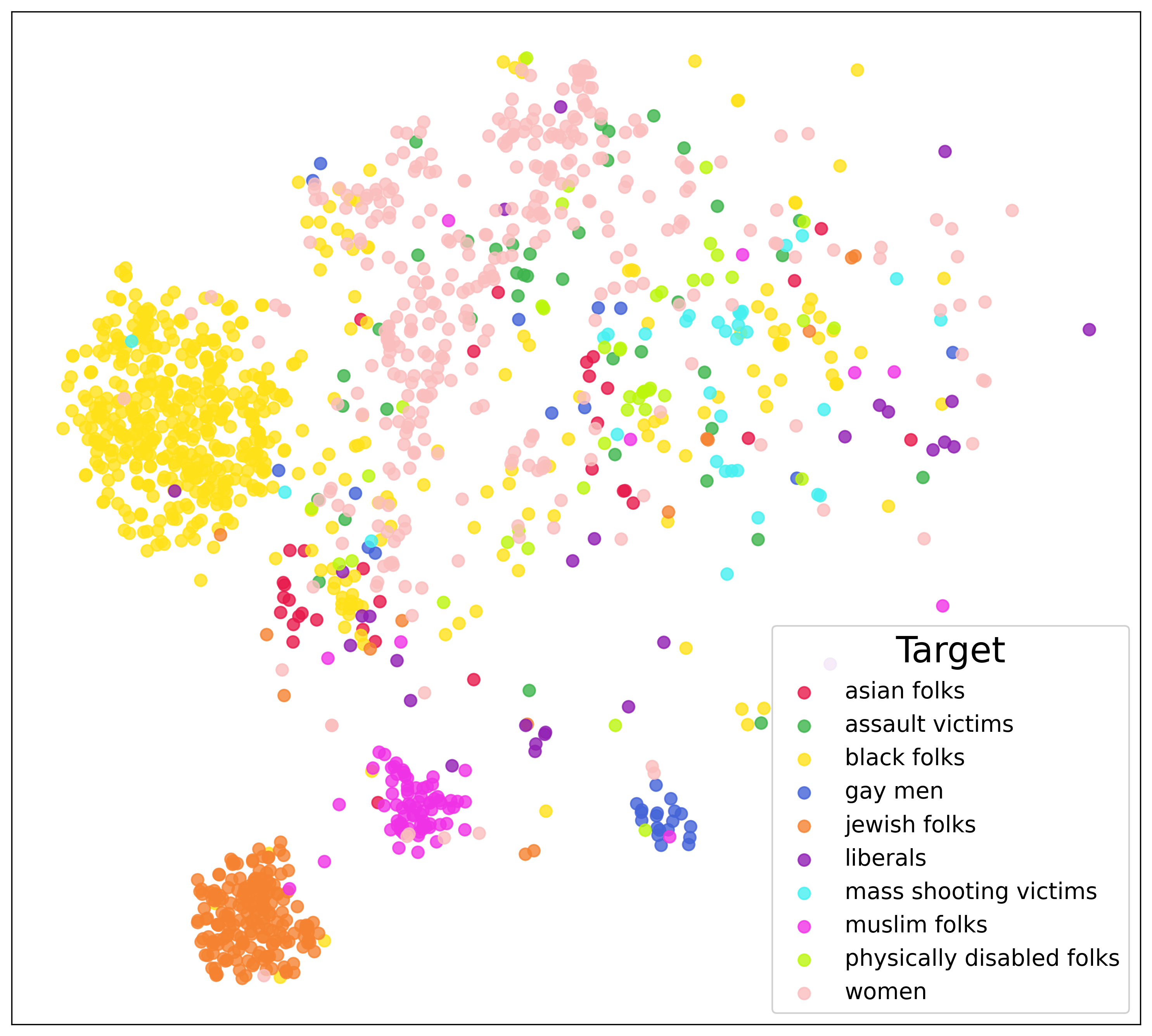}\caption{}\label{fig:qual_impsh_sbic_target}\end{subfigure} \\[0.3em]

\rotatebox{90}{\scriptsize\textsc{DynaHate}\hspace{0.5em}} &
\begin{subfigure}[b]{0.14\textwidth}\centering\includegraphics[width=\linewidth]{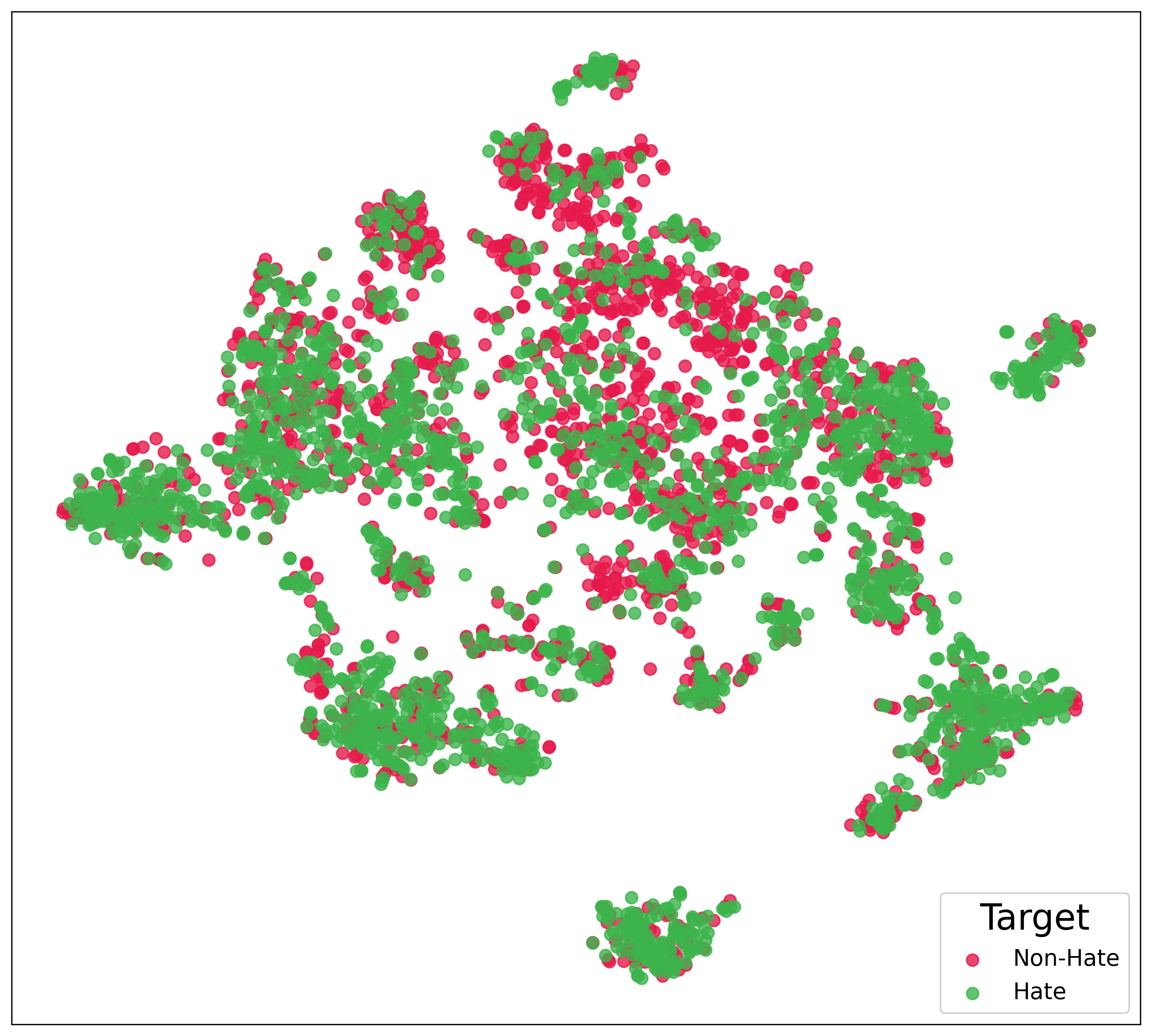}\caption{}\label{fig:qual_impcon_DynaHate_class}\end{subfigure} &
\begin{subfigure}[b]{0.14\textwidth}\centering\includegraphics[width=\linewidth]{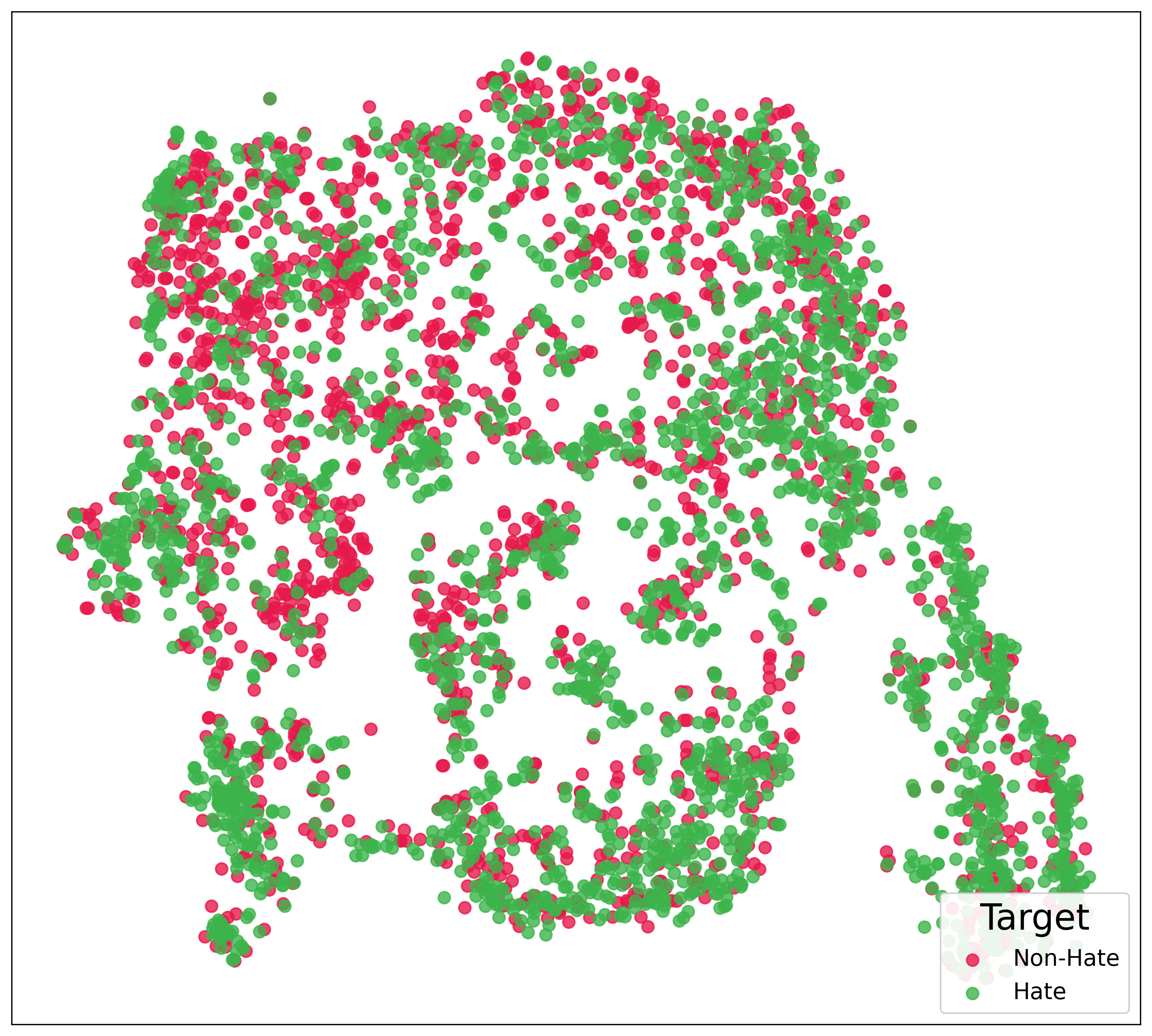}\caption{}\label{fig:qual_shared_DynaHate_class}\end{subfigure} &
\begin{subfigure}[b]{0.14\textwidth}\centering\includegraphics[width=\linewidth]{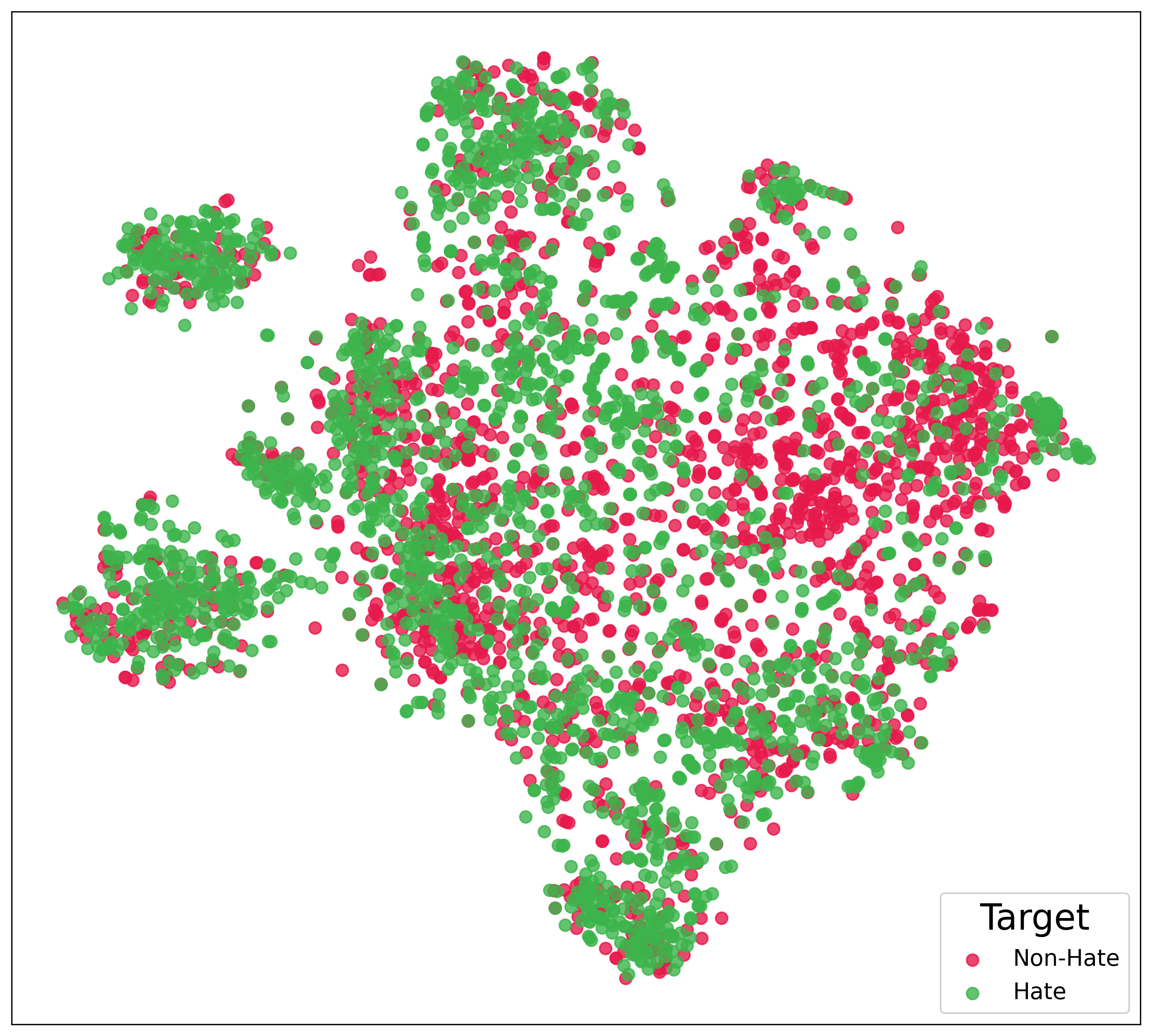}\caption{}\label{fig:qual_impsh_DynaHate_class}\end{subfigure} & & & \\
\end{tabular}
\caption{t-SNE visualization of test embeddings. Rows: \textsc{IHC} (in-domain), \textsc{SBIC} and \textsc{DynaHate} (cross-domain). Left: class-colored; right: target-colored. \textsc{ImpCon} fragments class boundaries into target-centric clusters. \textsc{SharedCon} shows partial separation. \textsc{ImpSH} maintains global Hate/Non-Hate separation across all domains.}
\label{fig:qual_ihc_tsne}
\end{figure*}

We further probe the embedding space with t\textsc{-}SNE projections that provide a qualitative view of how classes and target groups organize in the learned representation. We report a class-colored view and a target-colored view, where target labels serve as a rough proxy for shared topical content. We train on \textsc{IHC} as the source domain, visualize the \textsc{IHC} in-domain test split, and then visualize transfer to \textsc{SBIC} and \textsc{DynaHate}. On \textsc{IHC}, both \textsc{ImpCon} and \textsc{SharedCon} show target-consistent clusters in the target view (Fig.~\ref{fig:qual_ihc_tsne}d-f). In the class view, \textsc{ImpCon} shows more class interleaving within these clusters, while \textsc{SharedCon} and \textsc{ImpSH} exhibit a clearer large-scale separation between \textsc{Hate} and \textsc{Non-Hate} (Fig.~\ref{fig:qual_ihc_tsne}a-c). \textsc{ImpSH} appears to preserve target-coherent neighborhoods while reducing cross-class overlap in several regions of the map.

Under transfer to \textsc{SBIC}, class mixing increases for all methods, which is expected under a larger shift, but \textsc{ImpSH} shows comparatively less overlap in the class view while still maintaining target-coherent neighborhoods in the target view (Fig.~\ref{fig:qual_ihc_tsne}g-l). On \textsc{DynaHate}, all projections become more fragmented and mixed, yet \textsc{ImpSH} retains a weak but visible global class structure compared to the baselines (Fig.~\ref{fig:qual_ihc_tsne}m-o). These observations are qualitative and can vary with t\textsc{-}SNE settings, but they are broadly consistent with the Alignment and Uniformity trends in Table~\ref{tab:align}.

\section{Case Study}

Table~\ref{tab:alignment_tp_fn_k2} lists nearest neighbours from \textsc{ImpSH} with cosine similarity (lower is closer). True positives show that the model clusters coherent hate themes: on \textsc{SBIC}, misogynistic posts group together; on \textsc{DynaHate}, homophobic contexts align tightly. The false negatives expose three failure modes. First, topic bias: in \textsc{SBIC} the anchor is a non-hate “Stable Genius” joke, yet its neighbours are political Trump comments that match topic but not label, indicating target-centric proximity can override stance. Second, stance reversal: in \textsc{DynaHate} the anchor attacks immigrants while a neighbour condemns slurs, so opposite polarity is pulled close by shared lexical cues. Third, shallow lexical ties: “Rats are disgusting” is retrieved via short, high-overlap phrasing despite being off-topic. Overall, \textsc{ImpSH} groups most hate correctly at the sentence level, but residual errors arise from topic alignment dominating stance, brief or unequal contexts, and lexical shortcuts. 
\begin{table*}[!ht]
\centering
\small
\caption{\textbf{Alignment case study} ($k{=}2$). Rows show true‑positives (TP) and
false‑negatives (FN) for label hate (Offensive in \textsc{SBIC}, Hate in \textsc{DynaHate}). NN‑$i$ /
Sim‑$i$ give the $i$‑th neighbour and cosine similarity (lower = closer). }
\label{tab:alignment_tp_fn_k2}

\begin{tabular}{@{}p{0.12\linewidth}p{0.02\linewidth}p{0.02\linewidth}
    >{\raggedright\arraybackslash}p{0.19\linewidth}
    >{\raggedright\arraybackslash}p{0.19\linewidth}p{0.04\linewidth}
    >{\raggedright\arraybackslash}p{0.19\linewidth}p{0.04\linewidth}@{}}
\toprule
Dataset & Class & Outc & Post & NN‑1 & Sim‑1 & NN‑2 & Sim‑2 \\
\midrule
\textsc{SBIC} & off & TP &
...femcunt can’t last a week without sex, wants a superChad... &
...Women should be euthanized after menopause... & 0.68 &
...femoids give me disgusted looks in public... & 0.66 \\[4pt]

\textsc{SBIC} & off & FN &
...'Stable Genius' kept Trump rent‑free in NPR cat ladies' heads... &
...Trump never lets evidence stop his accusations... & 0.46 &
...Trump didn’t give me 100 \% of what I wanted on Jan 20... & 0.45 \\[4pt]

\textsc{DynaHate} & Hate & TP &
...Comcast router 69 fags aren’t as hidden as they think... &
...Met girl via app; angry hood, Portland feels third‑world... & 0.69 &
...Bearded guy at the bar makes me wary of being here... & 0.67 \\[4pt]

\textsc{DynaHate} & Hate& FN &
...holy sheeeeit that’s racist, immigrant rats... &
...'immigrants are disgusting' shouldn’t be feed& 0.51 &
Rats are disgusting& 0.50 \\
\bottomrule
\end{tabular}
\end{table*}

\section{Conclusion}
We proposed \textsc{ImpSH} and \textsc{AugSH}, two objectives that combine semantic priors with semi-hard negative mining to support transfer under domain shift in implicit hate detection. Across BERT and HateBERT, \textsc{ImpSH} is consistently competitive on cross-dataset evaluation and it is strongest in several transfer settings, most clearly when training on \textsc{IHC} with HateBERT. When training on \textsc{SBIC}, the best objective depends on the target and encoder, and clustering based baselines such as \textsc{SharedCon} or the augmentation ablation \textsc{AugSH} can be stronger in some cases. The geometry analysis supports this interpretation, where alignment and uniformity trends and qualitative visualizations suggest clearer Hate versus Non-Hate structure and less target specific fragmentation compared with prior contrastive objectives. Overall, our findings suggest that \textsc{ImpSH} facilitates superior separation in cross-data scenarios because it maps the input post somewhat bijectively to its underlying insinuation. By carefully mining negatives within the context of these implied statements, we mitigate the volatility of unsupervised clustering and achieve a more robust alignment of intent.

\section{Limitations and broader impact}
Our method inherits limitations from triplet learning with semi-hard mining. It uses a fixed margin, so training can be sensitive to the margin choice and to batch composition. Small batches can produce too few informative triplets and larger batches increase compute and memory due to within-batch pairwise distance computation. Empirically, gains over strong contrastive baselines are modest and not consistent across all transfer directions, and prior objectives can remain stronger for in-domain fitting, so we position \textsc{ImpSH} as a complementary objective rather than a replacement that dominates prior work \citep{kim2022generalizable,ahn2024sharedcon}. A further limitation is dependence on implication supervision because implied statements define positive pairs, which limits applicability when implication annotations are missing or noisy \citep{kim2022generalizable}. A practical extension is to replace the fixed margin with an adaptive margin scheme and to reduce reliance on implied statements by constructing positives through label-aware neighborhoods that cluster shared semantics within each label \citep{ahn2024sharedcon}, while still sampling negatives from nearby opposing clusters to keep the objective focused on borderline errors.

\section{Ethics statement}
This work proposes a training objective for hate related text classification. We evaluate on public datasets that contain hateful and abusive content, so readers may face secondary exposure. We include a content warning and we minimize qualitative examples. When examples are necessary, we keep excerpts short and we redact slurs and identity terms when this does not change the linguistic phenomenon under discussion.

Our objective uses semi-hard negative mining around borderline pairs. This focuses learning on difficult cases, but it can also amplify the impact of annotation noise and spurious cues such as identity markers that correlate with labels. For this reason, we emphasize cross-dataset evaluation and we recommend targeted error analysis that checks false positives for benign identity mentions and false negatives for implicit hate. Where metadata permits, we recommend reporting group-wise error rates and documenting dataset populations and labeling choices.

We do not present trained checkpoints as a ready to deploy moderation tool. Hate detection models can be misused for surveillance or censorship and they can produce harmful errors in deployment. If models or code are released, we recommend documenting intended use, failure modes, and evaluation conditions using established responsible NLP guidance and model documentation practices.

\bibliography{custom}
\appendix
\section{Seed Report}\label{sec:seed-rep}
Table~\ref{tab:seed_report} reports macro-F1 across four random seeds for in-domain and cross-domain evaluation. When trained on \textsc{IHC}, \textsc{BERT} ($\alpha{=}0.3$) achieves an in-domain mean macro-F1 of 0.783 (std 0.0025). Cross-domain evaluation yields 0.614 (std 0.0119) on \textsc{SBIC} and 0.592 (std 0.0036) on \textsc{DynaHate}. Under the same training set, \textsc{HateBERT} ($\alpha{=}0.4$) achieves 0.764 (std 0.0049) in-domain, and 0.650 (std 0.0148) on \textsc{SBIC} and 0.608 (std 0.0035) on \textsc{DynaHate} in cross-domain evaluation.When trained on \textsc{SBIC}, \textsc{BERT} ($\alpha{=}0.4$) achieves an in-domain mean macro-F1 of 0.841 (std 0.0022), with cross-domain results of 0.618 (std 0.0195) on \textsc{IHC} and 0.620 (std 0.0024) on \textsc{DynaHate}. \textsc{HateBERT} ($\alpha{=}0.2$) achieves 0.844 (std 0.0035) in-domain, with cross-domain results of 0.606 (std 0.0243) on \textsc{IHC} and 0.604 (std 0.0180) on \textsc{DynaHate}. Overall, the results are stable across seeds. The largest variance appears in the \textsc{SBIC}$\rightarrow$\textsc{IHC} transfer setting, suggesting higher sensitivity to initialization in that direction.

\begin{table*}[t]
\centering
\caption{Macro-F1 across four seeds on \textsc{ImpSh}. Seed/Mean are shown in \%, while Std is reported in raw F1 (0-1).}
\label{tab:seed_report}
\setlength{\tabcolsep}{6pt}
\begin{tabular}{lcllcccccc}
\toprule
\textbf{Model} & \textbf{$\alpha$} & \textbf{Train} & \textbf{Test} & \textbf{Seed 0} & \textbf{Seed 1} & \textbf{Seed 2} & \textbf{Seed 3} & \textbf{Mean} & \textbf{Std} \\
\midrule
\multirow{3}{*}{BERT}   & \multirow{3}{*}{0.3} & \multirow{3}{*}{IHC} & \textsc{IHC}   & 78.7\% & 78.3\% & 78.1\% & 78.3\% & 78.3\% & 0.0025 \\
             &           &            & \textsc{SBIC}   & 62.5\% & 61.3\% & 62.0\% & 59.7\% & 61.4\% & 0.0119 \\
             &           &            & \textsc{DynaHate} & 59.5\% & 59.2\% & 58.7\% & 59.4\% & 59.2\% & 0.0036 \\
\midrule
\multirow{3}{*}{HateBERT} & \multirow{3}{*}{0.4} & \multirow{3}{*}{IHC} & \textsc{IHC}   & 76.0\% & 76.2\% & 77.1\% & 76.2\% & 76.4\% & 0.0049 \\
             &           &            & \textsc{SBIC}   & 64.9\% & 63.0\% & 66.5\% & 65.5\% & 65.0\% & 0.0148 \\
             &           &            & \textsc{DynaHate} & 60.9\% & 60.5\% & 61.2\% & 60.5\% & 60.8\% & 0.0035 \\
\midrule
\multirow{3}{*}{BERT}   & \multirow{3}{*}{0.4} & \multirow{3}{*}{SBIC} & \textsc{SBIC}   & 84.0\% & 84.4\% & 83.8\% & 84.0\% & 84.1\% & 0.0022 \\
             &           &            & \textsc{IHC}   & 63.6\% & 59.2\% & 61.7\% & 62.9\% & 61.8\% & 0.0195 \\
             &           &            & \textsc{DynaHate} & 62.2\% & 61.8\% & 62.2\% & 61.9\% & 62.0\% & 0.0024 \\
\midrule
\multirow{3}{*}{HateBERT} & \multirow{3}{*}{0.2} & \multirow{3}{*}{SBIC} & \textsc{SBIC}   & 84.7\% & 84.5\% & 84.5\% & 83.9\% & 84.4\% & 0.0035 \\
             &           &            & \textsc{IHC}   & 62.7\% & 58.4\% & 58.6\% & 62.8\% & 60.6\% & 0.0243 \\
             &           &            & \textsc{DynaHate} & 62.9\% & 59.0\% & 59.2\% & 60.5\% & 60.4\% & 0.0180 \\
\bottomrule
\end{tabular}
\end{table*}

\section{Sampling Strategies Ablation}
To determine the optimal mining strategy for our triplet objective, we evaluate four sampling rules. These rules govern how negative examples are selected for each anchor-positive pair.

\begin{description}
  \item[\textbf{Random}] Select a same-class post at random. This ignores the implied statement for defining negatives.
  \item[\textbf{Hard Margin}] Use the implied statement as positive. Always pick the hardest negative that violates the margin.
  \item[\textbf{Semi-Hard}] Use the implied statement as positive. Choose a negative inside the margin but farther from the anchor than the positive.
  \item[\textbf{Semi-Hard + Fallback}] Attempt semi-hard first. If no suitable negative is found in the batch, fall back to the hard margin rule.
\end{description}

Our initial hypothesis was that the \textbf{Hard Margin} strategy would be most effective. By forcing the model to contend with the most confusing negative samples relative to a post's implied meaning, we expected it to learn to separate similar insinuations effectively.

However, the results in Table~\ref{tab:mining_strategy_appendix} point decisively to the superiority of \textbf{Semi-Hard} sampling for generalization. When trained on \textsc{IHC}, it achieves the highest F1 score on both cross-domain datasets. This strength is mirrored when training on \textsc{SBIC}, where Semi-Hard again secures the best performance on the \textsc{IHC} cross-domain task plus ties for the best score on \textsc{DynaHate}. Its robust performance, including tying for the best in-domain score, contrasts sharply with the volatility of other methods. Random sampling may perform well in-domain but fails on transfer, while Hard Margin is inconsistent across different setups.

This empirical outcome is consistent with the original analysis of triplet loss by \citep{schroff2015facenet}. They argue that focusing exclusively on the hardest negatives can yield unstable gradients plus lead to poor local minima early in training. The semi-hard approach provides more stable learning signals, balancing model improvement without risking collapse. Given its superior and consistent performance on cross-domain evaluation, we adopt the \textbf{Semi-Hard} sampling strategy for all main experiments.

\begin{table}[ht!]
\centering
 \setlength{\tabcolsep}{1.4pt} 
\caption{Macro F1 for each mining rule. Best score in each column is bold.}
\label{tab:mining_strategy_appendix}
\begin{tabular}{lccc}
\toprule
\textbf{Mining Strategy} & \textsc{IHC} & \textsc{SBIC} & \textsc{DynaHate} \\
\midrule
\multicolumn{4}{c}{Trained on \textsc{IHC}} \\ \midrule
Hard Margin     & 0.785     & 0.584     & 0.565 \\
Random        & \textbf{0.789} & 0.593     & 0.545 \\
Semi-Hard      & 0.787     & \textbf{0.625} & \textbf{0.595} \\
Semi-Hard + Fallback & 0.777     & 0.619     & 0.585 \\
\midrule
\multicolumn{4}{c}{Trained on \textsc{SBIC}} \\ \midrule
\textbf{Mining Strategy} & \textsc{SBIC} & \textsc{IHC} & \textsc{DynaHate} \\
\midrule
Hard Margin     & \textbf0.840 & 0.618     & 0.602 \\
Random        & 0.814     & 0.610     & 0.602 \\
Semi-Hard      & \textbf{0.847} & \textbf{0.627} & \textbf{0.629} \\
Semi-Hard + Fallback & 0.841     & 0.598     & 0.618 \\
\bottomrule
\end{tabular}
\end{table}

\paragraph{Implementation Details of Ablation}
We evaluate both BERT and HateBERT encoders on models trained with either \textsc{IHC} or \textsc{SBIC}. For the semi-hard triplet mining strategy, we swept the margin parameter \(\alpha \in \{0.1, 0.2, 0.3, 0.4, 0.5\}\) and selected the best-performing configuration per model-dataset pair based on validation performance. The optimal margins were: \(\alpha = 0.3\) for BERT trained on \textsc{IHC}, \(\alpha = 0.4\) for HateBERT trained on \textsc{IHC}, \(\alpha = 0.4\) for BERT trained on \textsc{SBIC}, and \(\alpha = 0.2\) for HateBERT trained on \textsc{SBIC}. we replicate only on seed 0.
\section{Alpha Study}
We conducted an $\alpha$ hyperparameter study using the same training setup and optimization settings described in Section~\ref{sec:implement}. 
Each configuration was evaluated by sweeping $\alpha \in \{0.1, 0.2, 0.3, 0.4, 0.5\}$, and performance was measured using macro-F1 averaged over multiple random seeds. 
The results for both in-domain and cross-domain evaluations are reported in Table~\ref{tab:alpha_study}. 
For each training setup, we select the $\alpha$ value that yields the best overall performance, which is highlighted in bold.

\begin{table*}[t]
 \centering
 \caption{Alpha study on \textsc{ImpSh}: macro-F1 (mean over seeds) for in-domain and cross-domain evaluation across different $\alpha$ values. Bold indicates the selected $\alpha$ per training setup (BERT+IHC: $\alpha{=}0.3$, HateBERT+IHC: $\alpha{=}0.4$, BERT+SBIC: $\alpha{=}0.4$, HateBERT+SBIC: $\alpha{=}0.2$).}
 \label{tab:alpha_study}
 \setlength{\tabcolsep}{6pt}
 \begin{tabular}{lllccccc}
 \toprule
 \textbf{Model} & \textbf{Train} & \textbf{Test} & \multicolumn{5}{c}{$\alpha$} \\
 \cmidrule(lr){4-8}
  & & & 0.1 & 0.2 & 0.3 & 0.4 & 0.5 \\
 \midrule

 \multirow{3}{*}{BERT} & \multirow{3}{*}{IHC} & \textsc{IHC} (in)   & 0.774 & 0.778 & \textbf{0.783} & 0.779 & 0.780 \\
            &           & \textsc{SBIC}     & 0.580 & 0.609 & \textbf{0.614} & 0.598 & 0.618 \\
            &           & \textsc{DynaHate}   & 0.563 & 0.577 & \textbf{0.592} & 0.586 & 0.584 \\
 \midrule

 \multirow{3}{*}{HateBERT} & \multirow{3}{*}{IHC} & \textsc{IHC} (in)  & 0.767 & 0.762 & 0.761 & \textbf{0.764} & 0.769 \\
              &           & \textsc{SBIC}    & 0.611 & 0.634 & 0.637 & \textbf{0.650} & 0.645 \\
              &           & \textsc{DynaHate}  & 0.566 & 0.589 & 0.606 & \textbf{0.608} & 0.606 \\
 \midrule

 \multirow{3}{*}{BERT} & \multirow{3}{*}{SBIC} & \textsc{SBIC} (in)   & 0.839 & 0.835 & 0.839 & \textbf{0.841} & 0.839 \\
            &            & \textsc{IHC}      & 0.602 & 0.609 & 0.612 & \textbf{0.618} & 0.613 \\
            &            & \textsc{DynaHate}   & 0.611 & 0.609 & 0.610 & \textbf{0.620} & 0.613 \\
 \midrule

 \multirow{3}{*}{HateBERT} & \multirow{3}{*}{SBIC} & \textsc{SBIC} (in) & 0.842 & \textbf{0.844} & 0.844 & 0.845 & 0.842 \\
              &            & \textsc{IHC}    & 0.600 & \textbf{0.606} & 0.604 & 0.603 & 0.594 \\
              &            & \textsc{DynaHate}  & 0.605 & \textbf{0.604} & 0.605 & 0.600 & 0.600 \\
 \bottomrule
 \end{tabular}
\end{table*}

\section{Representation quality through class alignment and uniformity}
To quantitatively evaluate the structure of the learned embedding space, we use the \textbf{Alignment} and \textbf{Uniformity} metrics proposed by \citet{wang2020understanding}. These metrics provide a formal way to measure two desirable properties of a representation: that similar samples should be mapped to nearby embeddings (intra-class compactness) and that dissimilar samples should be spread out evenly (inter-class separation). Embeddings are L2-normalized before computing distances.
\section*{Alignment}

Alignment measures the expected distance between embeddings of positive pairs. In our context, a positive pair consists of two samples belonging to the same class. A lower alignment score indicates that samples within the same class are more tightly clustered. It is defined as:

\begin{equation}
\mathcal{L}_{\text{align}}(f; r) \triangleq \underset{(x, y) \sim p_{\text{pos}}}{\mathbb{E}} [\|f(x) - f(y)\|_2^r]
\end{equation}

where $f$ is the encoder, $(x, y) \sim p_{\text{pos}}$ denotes a pair of samples drawn from the positive pair distribution (i.e., having the same class label), and $\alpha > 0$ is a parameter (typically set to 2).

\section*{Uniformity}

Uniformity measures how well the embeddings of negative pairs are distributed across the embedding space. A lower uniformity score indicates that dissimilar samples are spread further apart and more evenly, maximizing the entropy of the representation. It is defined as:

\begin{equation}
\mathcal{L}_{\text{uniform}}(f; t) \triangleq \log \underset{(x, y) \sim p_{\text{neg}}}{\mathbb{E}} [e^{-t\|f(x) - f(y)\|_2^2}]
\end{equation}

where $(x, y) \sim p_{\text{neg}}$ denotes a pair of samples drawn from the negative pair distribution (i.e., having different class labels), and $t > 0$ is a parameter (typically set to 2).

Together, these two metrics provide a comprehensive evaluation of the global structure of the embedding space, quantifying both the cohesion within classes and the separation between them.

\end{document}